\documentclass{article}
\usepackage{arxiv}
\usepackage[utf8]{inputenc}
\usepackage{natbib}
\usepackage[T1]{fontenc}
\usepackage{amsmath,amssymb,mathtools}
\usepackage{graphicx,booktabs,adjustbox,array,multirow}
\usepackage{algorithm,algpseudocode}
\usepackage{microtype}
\usepackage{xcolor}
\usepackage{placeins}
\usepackage{hyperref}
\usepackage{url}
\usepackage{enumitem}
\hypersetup{colorlinks=true,citecolor=teal,linkcolor=teal,urlcolor=teal,pdftitle={BIDETA: Brain-Inspired Data-Efficient Tactile Adaptation for Unseen Sensors},pdfauthor={Boheng Liu, Lan Wei, Ziyu Li, Chenghua Duan, Qing Li, Dandan Zhang, Xia Wu}}

\newcommand{\R}{\mathbb{R}}
\newcommand{\RowNorm}{\operatorname{RowNorm}}
\newcommand{\softmax}{\operatorname{softmax}}
\newcommand{\diag}{\operatorname{diag}}

\newcommand{\KL}{\operatorname{KL}} 
\newcommand{\NormPrior}{\mathcal{N}_{\pi}}
\title{BIDETA: Brain-Inspired Data-Efficient Tactile Adaptation for Unseen Sensors}
\author{
    Boheng Liu$^{1}$ \quad
    Lan Wei$^{2}$ \quad
    Ziyu Li$^{1}$\thanks{Corresponding authors: Ziyu Li (\texttt{ziyuli@bit.edu.cn})} \quad
    Chenghua Duan$^{1}$ \\
    \bfseries Qing Li$^{1}$ \quad
    Dandan Zhang$^{2}$ \quad
    Xia Wu$^{1}$ \\[0.5em]
    $^{1}$School of Computer Science and Technology, Beijing Institute of Technology \\
    Beijing, China \\
    $^{2}$Department of Bioengineering, Imperial College London, London, UK \\
    \texttt{boheng@bit.edu.cn}, \texttt{l.wei24@imperial.ac.uk}, \texttt{ziyuli@bit.edu.cn},
    \texttt{d.zhang17@imperial.ac.uk}
}
\renewcommand{\headeright}{Preprint}
\renewcommand{\undertitle}{Preprint}
\renewcommand{\shorttitle}{BIDETA: Brain-Inspired Data-Efficient Tactile Adaptation}
\date{}
\begin{document} 
\maketitle 
\begin{abstract} 

Vision-based tactile sensors provide high-resolution contact information for robotic perception and contact-rich manipulation, advancing embodied intelligence through more reliable physical interaction.
However, device-specific sensing mechanisms cause tactile foundation models to degrade on unfamiliar hardware. 
Existing cross-sensor methods often require calibration data, paired observations, or iterative training. 
To address this problem, we propose Brain-Inspired Data-Efficient Tactile Adaptation (BIDETA),  a gradient-free framework that uses a frozen tactile encoder and a few labeled target contacts to jointly predict labels for an unlabeled query batch. 
Inspired by the brain’s rapid sensory adaptation, BIDETA combines rapid support memory, support-conditioned spectral graphs, and reliability-gated recurrence to preserve pretrained representations, repair sensor-dependent feature neighborhoods, and integrate reliable cross-query evidence.
Experiments on SITR, TacVerse Shape, and TacQuad show that BIDETA substantially improves adaptation to unknown sensors: with only 10\% labeled target data on SITR, it raises mean Sparsh accuracy from 6.86\% for the frozen source classifier to 87.09\%, exceeding the strongest implemented prior comparison by 47.22 percentage points, and these gains generalize across datasets, pretrained backbones, and tactile tasks.
In the SITR timing benchmark with TVL, BIDETA also achieves approximately 20× faster target-sensor adaptation than the best baseline.
BIDETA thus offers a gradient-free, data-efficient route to deploying tactile models on new hardware.

\end{abstract} 
\section{Introduction}

With rapid progress in embodied intelligence, touch is becoming a core capability for agents that must perceive and understand the physical world through contact~\citep{luo2025tactile}. Vision-based tactile sensors convert contact-induced deformation of a soft skin into high-resolution images, giving robots access to contact geometry, force, and slip that external vision cannot observe directly~\citep{yuan2017gelsight,lambeta2020digit}. However, differences in optical design, elastomer mechanics, marker or pin structure, and field of view can cause the same contact to appear substantially different across sensors~\citep{wei2026tacverse,cong2026taceva}. Despite substantial advances in tactile foundation models, their performance often drops sharply on unseen sensors introduced by sensor replacement, hardware upgrades, or deployment across heterogeneous robot fleets. Rapid adaptation to unseen tactile sensors is therefore essential for advancing embodied intelligence, allowing existing tactile models to serve diverse hardware.

\begin{figure}[!t]
\centering
\includegraphics[width=\linewidth]{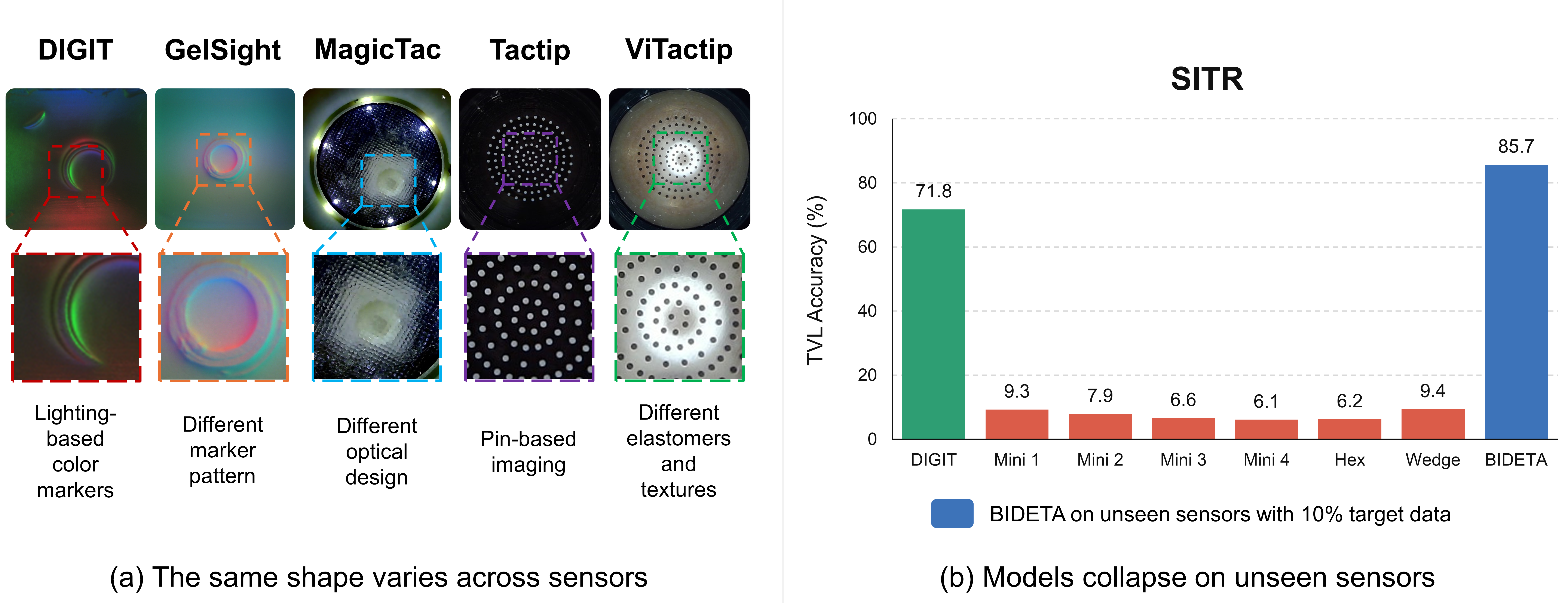}
\caption{\textbf{The same shape produces markedly different tactile observations across sensors, degrading model performance on unseen sensors.} (a) The same shape produces distinct tactile images across sensor designs. (b) A linear classifier trained on the source sensor is evaluated using frozen TVL features. On SITR~\citep{gupta2025sitr}, accuracy falls from 71.8\% on DIGIT to 9.3\%, 7.9\%, 6.6\%, 6.1\%, 6.2\%, and 9.4\% on Mini 1, Mini 2, Mini 3, Mini 4, Hex, and Wedge, respectively. With 10\% labeled target data, BIDETA reaches 85.7\% averaged across all six unknown sensors.}

\label{fig:motivation}
\end{figure}

Recent tactile foundation models reduce the cost of learning transferable features. TVL aligns touch with vision and language to learn semantically grounded tactile representations, while Sparsh uses self-supervised learning on heterogeneous tactile data to support transfer across downstream tasks~\citep{fu2024tvl,higuera2025sparsh}. However, the same contact can produce markedly different images across sensors, causing source-trained models to fail on unseen hardware. Figure~\ref{fig:motivation} shows this failure: a classifier trained with frozen TVL features performs well on its source sensor but falls close to chance on unknown sensors. Existing cross-sensor methods mitigate this shift, but often rely on calibration data, paired observations, or multi-sensor collection, limiting rapid adaptation to a new device~\citep{gupta2025sitr,feng2025anytouch,zhang2026ctsr}. Lighter-weight alternatives use prototypes, support caches, or query relations without updating the tactile encoder~\citep{wang2019simpleshot,zhang2022tipadapter,ziko2020laplacianshot}. Yet their effectiveness depends on pretrained feature geometry remaining meaningful after sensor shift; distorted target neighborhoods can make both support-based predictions and query relations unreliable. This exposes the central question of our work: \emph{how can a tactile foundation model use a few labeled target contacts to transfer its perceptual capability rapidly and reliably across unseen sensors?}

Even when pretrained tactile features retain useful class information, rapid and reliable transfer to unseen sensors remains challenging: sensor shift changes how classes are distinguished, distorts feature neighborhoods, and makes query relations unreliable. Data-efficient tactile adaptation should therefore satisfy three requirements. First, it should recover class evidence from a few labeled target contacts while preserving the pretrained representation. Second, it should use variation in those contacts to recalibrate feature neighborhoods, so that query relations remain meaningful on the new sensor. Third, it should determine how much relational evidence each query incorporates according to prediction reliability, without allowing unreliable relations to override support-derived evidence. These requirements call for a data-efficient method that rapidly adapts to unseen sensors with few labeled examples, without retraining the tactile foundation model.

To meet these requirements, we draw inspiration from the brain's rapid sensory adaptation mechanism~\citep{mcclelland1995cls,carandini2012normalization,ernst2002haptic,khona2022attractors} and propose Brain-Inspired Data-Efficient Tactile Adaptation (BIDETA). First, \emph{rapid support memory} fits discriminant memories to two readouts from the same frozen encoder; it then fuses them into a stable probability anchor. Second, \emph{support-conditioned spectral geometry} uses labeled target variation to suppress unstable feature directions; it then constructs a cross-readout consensus graph with more reliable neighborhoods. Third, \emph{reliability-gated anchored recurrence} uses prediction ambiguity and readout disagreement; these signals determine how strongly each query should absorb graph evidence while retaining its support-derived anchor. By separating stable memory from rapid geometric recalibration and reliability-controlled integration, BIDETA converts unknown-sensor transfer into a support-conditioned inference problem without gradient-based encoder updates, bridging the gap between pretrained sensor experience and previously unseen hardware.

Experiments across three tactile datasets, using two frozen backbones, demonstrate that BIDETA substantially improves perception on unknown sensors. With only 10\% labeled target data on SITR, BIDETA raises TVL accuracy from the frozen source classifier's 7.58\% to 85.71\% and exceeds the strongest external comparison by 38.24 percentage points. On TacVerse Shape, the same target-data budget raises Sparsh accuracy from 13.67\% to 83.58\%, exceeding the strongest external comparison by 27.60 percentage points. The gains extend across datasets, pretrained backbones, and tactile classification and retrieval tasks. These results show that brain-inspired rapid adaptation can strengthen cross-sensor tactile generalization, thereby offering a promising direction for robust embodied perception across evolving hardware.

\section{Related Work}
\textbf{Tactile foundation models.} Vision-based tactile learning began with sensors such as GelSight and DIGIT, which convert contact deformation into high-resolution images~\citep{yuan2017gelsight,lambeta2020digit}. Tactile foundation models progressed from early paired vision--touch learning with Touch and Go toward reusable multimodal and self-supervised representations~\citep{yang2022touchgo}. TVL aligns touch with vision and language; Sparsh learns transferable features from unlabeled tactile data; UniTouch and ViT-Lens connect heterogeneous sensors to shared multimodal representation spaces through sensor-specific tokens or pretrained visual encoders~\citep{fu2024tvl,higuera2025sparsh,yang2024unitouch,lei2024vitlens}. Recent systems further extend tactile learning to multisensory manipulation, distributed tactile skin, and visual--tactile material localization~\citep{higuera2025tacx,sharma2025percepskin,kim2026seeing}. These advances improve tactile perception, but the resulting models remain difficult to adapt efficiently to other sensors across diverse downstream tasks.

\textbf{Data-Efficient Adaptation to Unseen Tactile Sensors.} Data-efficient adaptation methods broadly follow optimization-based or relation-based strategies. Optimization-based methods fit a lightweight classifier, adapter, or test-time state from the support set, directly adjusting the decision boundary but requiring iterative updates and risking overfitting to scarce examples~\citep{karmanov2024tda,boudiaf2020tim,singh2026multimodality}. Relation-based methods retain frozen features and classify through normalized prototypes, support caches, probability objectives, or query graphs~\citep{wang2019simpleshot,zhang2022tipadapter,ziko2020laplacianshot,zhou2003consistency,martin2024transclip}. Data-efficient adaptation methods designed specifically for tactile hardware remain sparse. The closest cross-sensor methods use simulated variation and calibration in SITR, cross-sensor matching in AnyTouch, or synthetic transfer and sensor-conditioned modulation in CTSRL~\citep{gupta2025sitr,feng2025anytouch,zhang2026ctsr}. These methods require specific calibration or paired data, or struggle to adapt when an unseen sensor differs substantially from the training sensors. Such dependence on predefined acquisition protocols consequently limits rapid deployment with heterogeneous tactile hardware.

\textbf{Brain-inspired methods.} The human brain adapts efficiently to changing environments by combining specialized neural processes, flexible memory, and dynamic regulation, providing computational principles for more robust and efficient algorithm design~\citep{zador2023neuroai}. Recent brain-inspired methods model prefrontal component processes for language-model planning, cortico-hippocampal dual-memory circuits for continual learning, and adaptive excitation--inhibition balance for reservoir computing~\citep{webb2025planning,shi2025corticohippocampal,srinivasan2025balance}. By translating neural organization, memory, and self-regulation into computational mechanisms, these methods improve planning, continual adaptation, and efficiency while addressing the rigidity of uniformly structured learning systems under rapidly changing conditions and distribution shifts.

\section{Method}
\label{sec:method}

\subsection{Problem formulation}
We consider an unseen target tactile sensor with a labeled support set $\mathcal{S}=\{(x_i,y_i)\}_{i=1}^{n}$ and an unlabeled query batch $\mathcal{Q}=\{x_j\}_{j=1}^{m}$, where $x$ denotes a tactile observation, $y_i\in\{1,\ldots,C\}$ is its label, $C$ is the number of classes, and every class appears in $\mathcal{S}$. A pretrained encoder $f_\theta$ with fixed parameters $\theta$ provides two feature readouts $h^{(v)}(x)\in\R^{d_v}$, where $v\in\{1,2\}$ indexes the readout and $d_v$ is its dimension. Given $\mathcal{S}$ and $\mathcal{Q}$, our goal is to predict the $m$ query labels jointly without query annotations, source-data replay, or encoder updates.

\subsection{Brain-inspired design overview}
Human perception remains stable under changing sensory conditions, as the brain combines long-term knowledge with rapid sensory adaptation. Complementary learning protects established representations while incorporating new experience, sensory normalization recalibrates responses to current input statistics, and reliability-weighted recurrent integration accumulates uncertain evidence toward a stable interpretation~\citep{mcclelland1995cls,carandini2012normalization,ernst2002haptic,khona2022attractors}. Together, these processes allow perception to adjust quickly without discarding previously acquired knowledge.

\begin{figure}[!t]
\centering
\includegraphics[width=\linewidth]{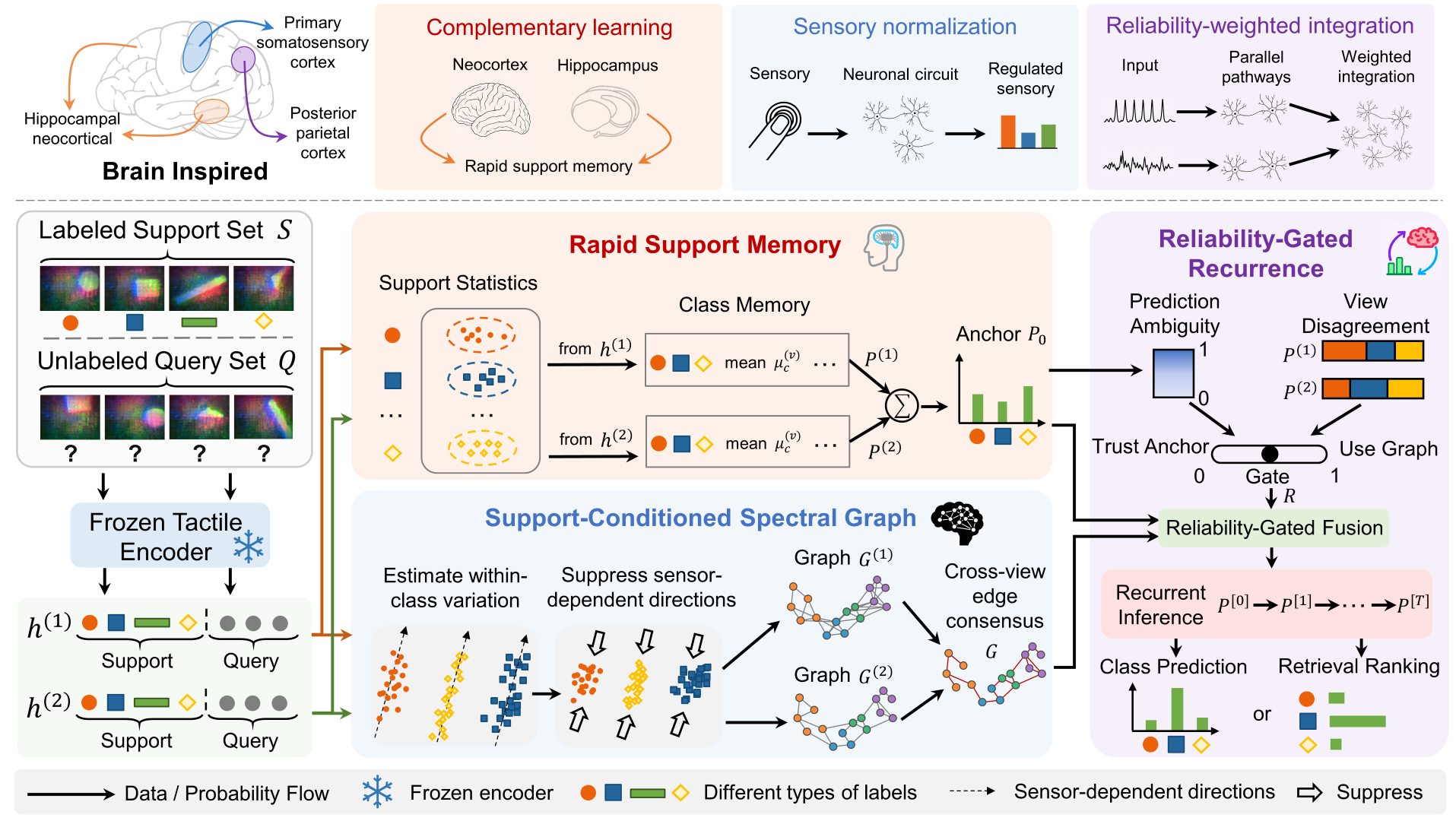}
\caption{\textbf{Information flow through BIDETA.} A frozen encoder produces two readouts. Rapid support memory forms the probability anchor $P_0$, while the support-conditioned spectral graph reshapes query geometry and constructs the cross-readout affinity matrix $G$. Prediction ambiguity and readout disagreement determine the query-wise reliability gate $R$. Anchored recurrent inference combines $P_0$, $G$, and $R$ to produce class probabilities or identity rankings.}
\label{fig:method}
\end{figure}

Inspired by this mechanism, BIDETA couples three modules, as shown in Figure~\ref{fig:method}. \emph{Rapid support memory} converts the two frozen readouts into a support-supervised probability anchor $P_0\in[0,1]^{m\times C}$. A \emph{support-conditioned spectral graph} suppresses unstable feature directions and produces a query affinity matrix $G\in\R_{+}^{m\times m}$, where $\R_{+}$ denotes the nonnegative real numbers. A \emph{reliability-gated recurrence} forms a diagonal matrix $R\in[0,1]^{m\times m}$ that controls how strongly each query uses relational evidence. Their joint action preserves stable class evidence while recalibrating sensor-dependent geometry and evidence flow for the unknown sensor.

\subsection{Rapid support memory}
\label{sec:memory}
The brain combines stable long-term representations with rapid learning from limited new experience, allowing perception to adapt without discarding prior knowledge. Inspired by this complementary learning principle, we design rapid support memory to preserve the frozen encoder while extracting class evidence from a small target support set.
For each readout $v$, we compute the support mean $b^{(v)}\in\R^{d_v}$ and coordinate-wise standard deviation $s^{(v)}\in\R^{d_v}$, then standardize every support or query feature as
\begin{equation}
 z^{(v)}(x)=\big(h^{(v)}(x)-b^{(v)}\big)\oslash s^{(v)},
 \label{eq:standardize}
\end{equation}
where $\oslash$ denotes coordinate-wise division and constant coordinates use unit scale. Let $n_c=|\{i:y_i=c\}|$, $\pi_c=n_c/n$, and $\mu_c^{(v)}$ denote the support count, support prior, and standardized mean of class $c$ in readout $v$. We fit a shrinkage linear discriminant analysis memory~\citep{fisher1936lda} with pooled within-class covariance $\Sigma_{\rho}^{(v)}\in\R^{d_v\times d_v}$ and shrinkage parameter $\rho\in[0,1]$. Its class score for a standardized feature $z$ is
\begin{equation}
 \ell_c^{(v)}(z)=
 z^\top(\Sigma_{\rho}^{(v)})^{-1}\mu_c^{(v)}
 -\frac{1}{2}(\mu_c^{(v)})^\top(\Sigma_{\rho}^{(v)})^{-1}\mu_c^{(v)}
 +\log\pi_c .
 \label{eq:lda}
\end{equation}
For query $x_j$, a temperature $\tau_m>0$ converts the scores into the class distribution $P_{j}^{(v)}\in[0,1]^C$ through
\begin{equation}
 P_{jc}^{(v)}=\softmax_c\!\left(\ell_c^{(v)}(z^{(v)}(x_j))/\tau_m\right),
 \qquad
 P_0=\alpha P^{(1)}+(1-\alpha)P^{(2)},
 \label{eq:anchor}
\end{equation}
where $\softmax_c$ normalizes over the $C$ class scores, $\alpha\in[0,1]$ is the readout weight, and $P_0$ is the fused probability anchor. This anchor carries direct support-supervised class evidence into the final inference stage.

\subsection{Support-conditioned spectral graph}
\label{sec:spectral}
Neural sensory systems normalize responses according to current input statistics, reducing the influence of dominant but uninformative variation. Drawing on this adaptive regulation principle, we design the support-conditioned spectral graph to use within-class target variation to reshape feature geometry and obtain more reliable query neighborhoods.
Sensor changes can enlarge feature directions that vary within a class, causing them to dominate query similarity. For each support example, we define the class residual $e_i^{(v)}=z^{(v)}(x_i)-\mu_{y_i}^{(v)}$ and estimate its covariance with Ledoit--Wolf shrinkage~\citep{ledoit2004covariance}:
\begin{equation}
 \widehat\Sigma_w^{(v)}
 =\operatorname{LW}\!\left(\{e_i^{(v)}\}_{i=1}^{n}\right)
 =U^{(v)}\diag(\lambda_1^{(v)},\ldots,\lambda_{d_v}^{(v)})(U^{(v)})^\top ,
 \label{eq:cov}
\end{equation}
where $U^{(v)}$ contains the eigenvectors and $\lambda_a^{(v)}$ is the eigenvalue of direction $a$. Given a spectral exponent $\gamma\geq0$ and numerical floor $\epsilon=10^{-6}$, we downweight high-variance directions through
\begin{align}
 g_a^{(v)}
 &=\frac{\max(\lambda_a^{(v)},\epsilon)^{-\gamma}}
 {\operatorname{median}_{b}\max(\lambda_b^{(v)},\epsilon)^{-\gamma}},
 &A^{(v)}&=U^{(v)}\diag(g^{(v)})(U^{(v)})^\top,\nonumber\\
 \widetilde z_j^{(v)}&=A^{(v)}z^{(v)}(x_j),
 \label{eq:spectral}
\end{align}
where $g_a^{(v)}$ is the gain for direction $a$, $A^{(v)}$ is the spectral transform, and $\widetilde z_j^{(v)}$ is the transformed query feature. The transform changes only the geometry used to connect queries and leaves the probability anchor in Eq.~\eqref{eq:anchor} unchanged.

We normalize $q_j^{(v)}=\widetilde z_j^{(v)}/\|\widetilde z_j^{(v)}\|_2$ and construct a directed $k$-nearest-neighbor graph. Let $\mathcal{N}_k^{(v)}(j)$ be the neighbors of query $j$ in readout $v$, excluding $j$, and let $\tau_g>0$ be the graph temperature. The directed edge from query $j$ to query $l$ is
\begin{equation}
 W_{jl}^{(v)}
 =\frac{\mathbf{1}[l\in\mathcal{N}_k^{(v)}(j)]
 \exp((q_j^{(v)})^\top q_l^{(v)}/\tau_g)}
 {\sum_{a\in\mathcal{N}_k^{(v)}(j)}
 \exp((q_j^{(v)})^\top q_a^{(v)}/\tau_g)} .
 \label{eq:knn}
\end{equation}
Here $\mathbf{1}[\cdot]$ is the indicator function. We symmetrize each readout graph and retain relations supported by both readouts:
\begin{equation}
 G^{(v)}=\RowNorm\!\left(\frac{W^{(v)}+(W^{(v)})^\top}{2}\right),
 \qquad
 G=\RowNorm\!\left(\sqrt{G^{(1)}\odot G^{(2)}}\right),
 \label{eq:consensus}
\end{equation}
where $\RowNorm$ divides each nonzero row by its sum, $\odot$ is entry-wise multiplication, and the square root is entry-wise. If a query has no shared edge, its row is replaced by the corresponding row of $(G^{(1)}+G^{(2)})/2$ before normalization. The resulting row-stochastic matrix $G$ encodes the support-adapted query geometry.

\subsection{Reliability-gated recurrence}
\label{sec:gate}
The brain integrates multiple sensory cues according to their confidence, relying more on contextual evidence when the current percept is uncertain. Motivated by this reliability-aware integration, we design the reliability gate to use prediction ambiguity and cross-readout disagreement to regulate each query's access to graph evidence.

The gate measures whether each query has a stable initial prediction. For class distributions $u,v\in[0,1]^C$, define entropy $H(u)=-\sum_{c=1}^{C}u_c\log u_c$ and divergence $\KL(u\|v)=\sum_{c=1}^{C}u_c\log(u_c/v_c)$. For query $j$, let $P_{0,j}$ be row $j$ of $P_0$ and $\overline P_j=(P_j^{(1)}+P_j^{(2)})/2$. We measure anchor ambiguity $a_j\in[0,1]$ and cross-readout disagreement $d_j\in[0,1]$ as~\citep{shannon1948entropy,lin1991js}
\begin{equation}
 a_j=\frac{H(P_{0,j})}{\log C},
 \qquad
 d_j=\sqrt{\frac{
 \tfrac12\KL(P_j^{(1)}\|\overline P_j)
 +\tfrac12\KL(P_j^{(2)}\|\overline P_j)}
 {\log 2}} .
 \label{eq:reliability}
\end{equation}
Given a disagreement weight $\lambda\in[0,1]$, gate exponent $p>0$, and recurrence bounds $0\leq r_{\min}\leq r_{\max}<1$, the combined uncertainty $s_j$ and recurrence weight $r_j$ are
\begin{equation}
 s_j=\operatorname{clip}\!\left((1-\lambda)a_j+\lambda d_j,0,1\right),
 \qquad
 r_j=r_{\min}+(r_{\max}-r_{\min})s_j^p,
 \qquad
 R=\diag(r_1,\ldots,r_m).
 \label{eq:gate}
\end{equation}
Here $\operatorname{clip}(u,0,1)$ truncates $u$ to $[0,1]$. A larger $r_j$ gives query $j$ greater access to graph evidence, while a smaller value keeps its prediction closer to the support-derived anchor.

Perceptual decisions are gradually refined through the interaction between retained knowledge and newly accumulated sensory evidence. Inspired by this recurrent integration process, BIDETA iteratively combines the stable support-derived anchor with reliability-controlled graph information, refining query predictions while retaining their initial class evidence.
Starting from $P^{[0]}=P_0$, BIDETA combines the anchor, spectral graph, and reliability gate for $T\in\mathbb{N}_{+}$ iterations:
\begin{equation}
 P^{[t+1]}=\NormPrior\!\left[(I_m-R)P_0+RGP^{[t]}\right],
 \qquad t=0,\ldots,T-1,
 \label{eq:update}
\end{equation}
where $P^{[t]}\in[0,1]^{m\times C}$ is the query probability matrix at iteration $t$ and $I_m$ is the $m\times m$ identity matrix. The first term restores support-supervised evidence, while the second propagates neighborhood evidence in proportion to each query's reliability weight.

The operator $\NormPrior$ clips its nonnegative input below $10^{-8}$ and applies five alternating column and row normalizations~\citep{cuturi2013sinkhorn}. For an intermediate matrix $V\in\R_{+}^{m\times C}$, each normalization step is
\begin{equation}
 V_{jc}\leftarrow V_{jc}\frac{m\pi_c}{\sum_{l=1}^{m}V_{lc}+10^{-12}},
 \qquad
 V_{jc}\leftarrow\frac{V_{jc}}{\sum_{b=1}^{C}V_{jb}+10^{-12}} .
 \label{eq:prior}
\end{equation}
This operation aligns aggregate query mass with the support prior $\pi=(\pi_1,\ldots,\pi_C)$ while returning each row to a class distribution. After $T$ iterations, classification outputs $\widehat y_j=\arg\max_c P_{jc}^{[T]}$, whereas retrieval ranks enrolled identities by sorting $P_{jc}^{[T]}$ over $c$ in descending order. Appendix~\ref{app:algorithm} provides the complete pseudocode.
 
\section{Experiments}
\label{sec:experiments}
 
\subsection{Experimental setup}
\paragraph{Datasets and label budgets.}
SITR~\citep{gupta2025sitr} contains seven sensors and 16 household-object classes. We use DIGIT as the source and four GelSight Mini devices, GelSight Hex, and GelSight Wedge as six targets; all are unseen by TVL, while Hex and Wedge are unseen sensor families for Sparsh. From 800 training images per target class, 1\%, 5\%, and 10\% budgets provide 8, 40, and 80 support images, and the 200-image evaluation split gives 3,200 queries per target. TacVerse Shape~\citep{wei2026tacverse} contains seven sensors and nine shape classes. We use GelSightNoMarker as the source and MagicGripper, MagicTac, TacTip, ViTac, and ViTacTip as five targets, all unseen by both backbones. Each sensor--class pair has 300 training, 100 validation, and 100 test images; the three budgets sample 3, 15, and 30 support images per class and use 900 test queries per target. TacQuad~\citep{feng2025anytouch} contains 56 enrolled trial identities, each a 20-frame object-contact sequence, across three RGB sensors. DIGIT is the source, while GelSight Mini and DuraGel are two targets; both are unseen by TVL, and DuraGel is unseen by Sparsh. We sample 2, 4, or 6 support frames per identity at 10\%, 20\%, and 30\%, reserve four frames as a temporal gap, and use the final four frames to form 224 queries per target. Appendix~\ref{app:data} lists the label spaces and exact splits.

\paragraph{Encoders and sensor exposure.}
Both TVL ViT-Small and Sparsh-DINO Small remain frozen~\citep{fu2024tvl,higuera2025sparsh}. Their architectures follow the Vision Transformer~\citep{dosovitskiy2021vit}, and Sparsh uses DINO self-distillation~\citep{caron2021dino}. TVL takes $224\times224$ RGB images, whereas Sparsh concatenates two RGB frames along the channel dimension. SITR and TacVerse duplicate the same frame for Sparsh, while TacQuad pairs each frame with an earlier neighboring frame from the same trial. TVL was pretrained with DIGIT; Sparsh's pretraining includes DIGIT, GelSight 2017, and GelSight Mini. Appendix~\ref{app:preprocessing} details preprocessing.

\begin{table}[!htbp]
\centering
\caption{\textbf{Data-efficient classification on six unseen SITR sensors with two pretrained backbones under different label budgets.} Accuracy (\%) is averaged across the unknown sensors and reported as mean $\pm$ sample standard deviation over three random seeds. I/T denotes inductive/transductive inference. Best results are in bold and second-best results are underlined.}
\label{tab:sitr}

\setlength{\tabcolsep}{3pt}
\begin{adjustbox}{max width=\linewidth}
\begin{tabular}{lccccccc}
\toprule
Method & Mode & TVL 1\% & TVL 5\% & TVL 10\% & Sparsh 1\% & Sparsh 5\% & Sparsh 10\% \\
\midrule
Frozen backbone & I & $7.58 \pm 0.14$ & $7.58 \pm 0.14$ & $7.58 \pm 0.14$ & $6.86 \pm 0.33$ & $6.86 \pm 0.33$ & $6.86 \pm 0.33$ \\
Tip-Adapter & I & $17.21 \pm 1.63$ & $23.51 \pm 0.51$ & $28.56 \pm 1.37$ & $14.85 \pm 0.34$ & $21.90 \pm 0.92$ & $22.30 \pm 0.04$ \\
SimpleShot & I & $37.98 \pm 0.69$ & $44.47 \pm 0.90$ & $44.55 \pm 0.62$ & $32.08 \pm 2.45$ & $38.64 \pm 0.96$ & $39.21 \pm 0.81$ \\
LaplacianShot & T & $\underline{38.55 \pm 0.38}$ & $45.84 \pm 1.65$ & $44.11 \pm 0.60$ & $\underline{32.78 \pm 3.52}$ & $\underline{39.94 \pm 1.29}$ & $\underline{39.87 \pm 0.77}$ \\
SITR-Calib  & I & $38.00 \pm 1.56$ & $\underline{46.58 \pm 0.08}$ & $\underline{47.47 \pm 1.44}$ & $30.88 \pm 1.99$ & $34.46 \pm 1.00$ & $34.17 \pm 0.35$ \\
AnyTouch Match  & I & $36.27 \pm 1.55$ & $44.30 \pm 0.12$ & $45.32 \pm 1.00$ & $32.34 \pm 0.99$ & $35.36 \pm 0.56$ & $35.81 \pm 0.64$ \\
CTSRL CSM  & I & $37.42 \pm 1.77$ & $43.87 \pm 1.29$ & $44.99 \pm 0.85$ & $31.98 \pm 1.27$ & $34.94 \pm 0.25$ & $34.31 \pm 0.39$ \\
BIDETA & T & $\mathbf{62.22 \pm 1.11}$ & $\mathbf{83.59 \pm 0.81}$ & $\mathbf{85.71 \pm 0.12}$ & $\mathbf{62.12 \pm 1.43}$ & $\mathbf{83.81 \pm 0.80}$ & $\mathbf{87.09 \pm 0.61}$ \\
\bottomrule
\end{tabular} 
\end{adjustbox}
\end{table}

\begin{table}[!htbp]
\centering
\caption{\textbf{Data-efficient classification on five unseen TacVerse Shape sensors with two pretrained backbones under different label budgets.} Accuracy (\%) is averaged across the unknown sensors and reported as mean $\pm$ sample standard deviation over three random seeds. I/T denotes inductive/transductive inference. SITR-Calib is unavailable because standard calibration images are absent. Best results are in bold and second-best results are underlined.}
\label{tab:tacverse}

\setlength{\tabcolsep}{3pt}
\begin{adjustbox}{max width=\linewidth}
\begin{tabular}{lccccccc}
\toprule
Method & Mode & TVL 1\% & TVL 5\% & TVL 10\% & Sparsh 1\% & Sparsh 5\% & Sparsh 10\% \\
\midrule
Frozen backbone & I & $9.56 \pm 0.48$ & $9.56 \pm 0.48$ & $9.56 \pm 0.48$ & $13.67 \pm 0.47$ & $13.67 \pm 0.47$ & $13.67 \pm 0.47$ \\
Tip-Adapter & I & $22.79 \pm 4.24$ & $39.45 \pm 2.15$ & $45.52 \pm 0.73$ & $14.24 \pm 1.07$ & $25.61 \pm 0.74$ & $35.43 \pm 0.72$ \\
SimpleShot & I & $\underline{44.57 \pm 2.70}$ & $51.13 \pm 1.73$ & $53.35 \pm 0.76$ & $\underline{46.61 \pm 0.54}$ & $\underline{53.80 \pm 0.91}$ & $\underline{55.98 \pm 0.86}$ \\
LaplacianShot & T & $43.57 \pm 4.28$ & $49.66 \pm 1.61$ & $51.61 \pm 1.26$ & $44.96 \pm 2.31$ & $51.47 \pm 0.60$ & $52.24 \pm 1.17$ \\
SITR-Calib  & I & -- & -- & -- & -- & -- & -- \\
AnyTouch Match  & I & $42.44 \pm 1.93$ & $51.61 \pm 1.73$ & $55.83 \pm 0.37$ & $41.33 \pm 1.71$ & $49.23 \pm 1.16$ & $49.82 \pm 2.12$ \\
CTSRL CSM  & I & $43.26 \pm 1.41$ & $\underline{53.61 \pm 3.41}$ & $\underline{56.67 \pm 0.83}$ & $42.44 \pm 1.93$ & $49.50 \pm 1.82$ & $49.74 \pm 1.97$ \\
BIDETA & T & $\mathbf{57.93 \pm 1.73}$ & $\mathbf{75.47 \pm 2.03}$ & $\mathbf{79.13 \pm 1.20}$ & $\mathbf{59.66 \pm 2.61}$ & $\mathbf{80.65 \pm 0.43}$ & $\mathbf{83.58 \pm 1.05}$ \\
\bottomrule
\end{tabular}
\end{adjustbox}
\end{table}

\paragraph{Comparison methods.}
We use Tip-Adapter~\citep{zhang2022tipadapter}, SimpleShot~\citep{wang2019simpleshot}, and LaplacianShot~\citep{ziko2020laplacianshot} as representative frozen-feature data-efficient methods. They transfer support information through cache affinity, normalized class prototypes, or transductive query relations without updating the tactile encoder. We further include three recent tactile cross-sensor adaptation methods: SITR-Calib~\citep{gupta2025sitr}, AnyTouch Match~\citep{feng2025anytouch}, and CTSRL CSM~\citep{zhang2026ctsr}. 
These methods retain frozen features while training a lightweight feature module and classification head. Appendix~\ref{app:baselines} gives their implementations.

\paragraph{Metrics and experimental settings.}
Classification uses accuracy and macro-F1~\citep{sokolova2009metrics}; ranking uses mean reciprocal rank (MRR)~\citep{craswell2009mrr} and recall at one (R@1), following standard ranked-retrieval evaluation~\citep{manning2008ir}. We report the mean and sample standard deviation over three random seeds. All experiments run on an NVIDIA H800 GPU with 80\,GB memory. Appendix~\ref{app:settings} reports the complete implementation and parameters.

\subsection{Cross-sensor classification}

Table~\ref{tab:sitr} shows that BIDETA improves unseen-sensor performance at every backbone--budget setting on SITR. With TVL and only 1\% labeled target data, BIDETA raises accuracy from 7.58\% for the frozen source classifier to 62.22\%, outperforming the strongest comparison by 23.67 percentage points. At the 10\% budget, it raises accuracy from 7.58\% to 85.71\% and exceeds the strongest comparison by 38.24 percentage points. With Sparsh, BIDETA reaches 62.12\%, 83.81\%, and 87.09\% at the three budgets, outperforming LaplacianShot by 29.34, 43.87, and 47.22 percentage points, respectively.
Table~\ref{tab:tacverse} evaluates whether BIDETA's performance advantage transfers to another dataset. With TVL at the 10\% budget, BIDETA reaches 79.13\% accuracy and outperforms the strongest comparison by 22.46 percentage points. With Sparsh, BIDETA achieves 83.58\% and improves over the strongest comparison by 27.60 percentage points. These results demonstrate that BIDETA robustly generalizes across datasets and improves unknown-sensor recognition with different pretrained backbones.

\begin{figure}[!t]
\centering
\includegraphics[width=\linewidth]{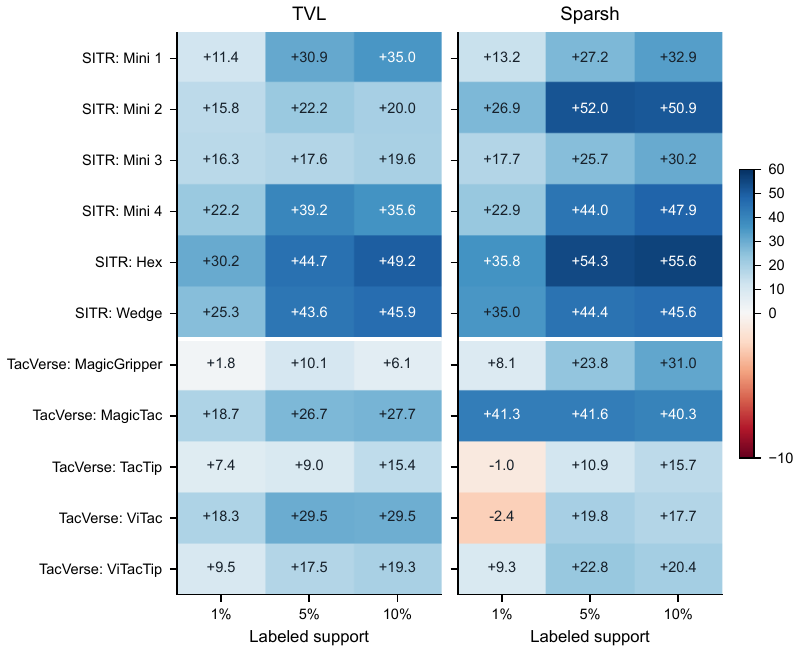}
\caption{\textbf{Performance comparison with the strongest baseline across sensors on SITR and TacVerse.} Each cell reports the accuracy gain of BIDETA for the corresponding unseen sensor and label budget.}
\label{fig:sensorgain}
\end{figure}

Figure~\ref{fig:sensorgain} reports BIDETA's gain over the strongest comparison for every unseen sensor on SITR and TacVerse with both pretrained backbones. The largest gains occur on GelSight Hex in SITR and MagicTac in TacVerse, reaching 55.6 and 41.6 percentage points, respectively. 
On TacVerse, negative gains occur only with Sparsh on TacTip and ViTac at the 1\% budget. We suspect that the three randomly sampled examples per class may sometimes be unrepresentative, leaving insufficient target variation for BIDETA to reliably recalibrate feature geometry. Additional sensor-wise, class-wise, 2-shot, 5-shot, 10-shot and macro-F1 results are reported in Appendix~\ref{app:classification}. 
\FloatBarrier

\begin{table}[!htbp]
\centering
\caption{\textbf{MRR results for closed-set identity ranking on TacQuad.} MRR is reported as a percentage, with mean $\pm$ sample standard deviation over three random seeds. SITR-Support uses the support--source mean difference as its condition. I/T denotes inductive/transductive inference. Best results are in bold and second-best results are underlined.}
\label{tab:tacquad}

\setlength{\tabcolsep}{3pt}
\begin{adjustbox}{max width=\linewidth}
\begin{tabular}{lccccccc}
\toprule
Method & Mode & TVL 10\% & TVL 20\% & TVL 30\% & Sparsh 10\% & Sparsh 20\% & Sparsh 30\% \\
\midrule
Frozen backbone & I & $8.76 \pm 0.00$ & $8.76 \pm 0.00$ & $8.76 \pm 0.00$ & $10.01 \pm 0.00$ & $10.01 \pm 0.00$ & $10.01 \pm 0.00$ \\
Tip-Adapter & I & $36.68 \pm 3.70$ & $47.80 \pm 12.36$ & $53.80 \pm 9.55$ & $45.80 \pm 0.69$ & $61.42 \pm 1.17$ & $64.57 \pm 0.73$ \\
SimpleShot & I & $\underline{59.05 \pm 0.78}$ & $64.65 \pm 1.65$ & $66.57 \pm 1.06$ & $60.36 \pm 0.85$ & $65.83 \pm 1.20$ & $68.75 \pm 0.78$ \\
LaplacianShot & T & $45.11 \pm 3.03$ & $48.78 \pm 2.27$ & $47.64 \pm 1.44$ & $43.32 \pm 3.10$ & $43.61 \pm 2.61$ & $45.82 \pm 0.88$ \\
SITR-Support & I & $57.94 \pm 2.00$ & $64.54 \pm 0.38$ & $65.49 \pm 1.42$ & $61.49 \pm 1.62$ & $67.90 \pm 0.63$ & $70.72 \pm 0.44$ \\
AnyTouch Match & I & $58.29 \pm 1.37$ & $\underline{66.15 \pm 2.49}$ & $\underline{68.49 \pm 1.10}$ & $\underline{61.73 \pm 0.93}$ & $\underline{67.93 \pm 0.85}$ & $\underline{71.00 \pm 0.51}$ \\
CTSRL CSM & I & $57.49 \pm 2.16$ & $63.88 \pm 0.88$ & $65.38 \pm 1.35$ & $58.88 \pm 0.78$ & $64.99 \pm 1.11$ & $68.14 \pm 0.23$ \\
BIDETA & T & $\mathbf{61.16 \pm 0.95}$ & $\mathbf{76.98 \pm 0.20}$ & $\mathbf{80.90 \pm 1.89}$ & $\mathbf{69.28 \pm 0.72}$ & $\mathbf{80.82 \pm 2.67}$ & $\mathbf{85.86 \pm 1.80}$ \\
\bottomrule
\end{tabular}
\end{adjustbox}
\end{table} 
\FloatBarrier

\begin{table}[!htbp]
\centering
\caption{\textbf{Ablation results on SITR.} Accuracy (\%) is reported as mean $\pm$ sample standard deviation over three random seeds. Best results are in bold and second-best results are underlined.}
\label{tab:ablation}

\setlength{\tabcolsep}{3pt}
\begin{adjustbox}{max width=\linewidth}
\begin{tabular}{lcccccc}
\toprule
Variant & TVL 1\% & TVL 5\% & TVL 10\% & Sparsh 1\% & Sparsh 5\% & Sparsh 10\% \\
\midrule
\multicolumn{7}{l}{\textit{Feature-memory ablations}} \\
Memory: view 1 only & $50.37 \pm 0.85$ & $72.49 \pm 0.99$ & $75.85 \pm 0.32$ & $51.65 \pm 3.06$ & $75.13 \pm 0.23$ & $77.65 \pm 0.59$ \\
Memory: view 2 only & $49.84 \pm 1.67$ & $70.82 \pm 1.01$ & $76.00 \pm 0.15$ & $54.54 \pm 1.38$ & $71.38 \pm 0.20$ & $77.62 \pm 0.82$ \\
Dual memory; no graph & $51.71 \pm 0.88$ & $72.74 \pm 1.02$ & $76.07 \pm 0.23$ & $52.88 \pm 2.88$ & $74.56 \pm 0.20$ & $77.92 \pm 0.80$ \\
\midrule
\multicolumn{7}{l}{\textit{Graph-geometry ablations}} \\
+ Consensus recurrence & $56.42 \pm 1.00$ & $77.41 \pm 0.57$ & $80.10 \pm 0.21$ & $55.53 \pm 2.57$ & $78.62 \pm 0.63$ & $80.86 \pm 0.67$ \\
+ Support standardization & $57.01 \pm 1.02$ & $77.99 \pm 0.58$ & $80.49 \pm 0.21$ & $56.18 \pm 2.50$ & $79.19 \pm 0.63$ & $81.51 \pm 0.63$ \\
+ Spectral normalization & $58.73 \pm 1.11$ & $81.48 \pm 0.77$ & $82.68 \pm 0.11$ & $59.17 \pm 2.25$ & $81.70 \pm 0.90$ & $84.14 \pm 0.83$ \\
Full: arithmetic graph fusion & $61.76 \pm 1.17$ & $83.24 \pm 0.79$ & $\underline{85.70 \pm 0.15}$ & $\underline{61.89 \pm 1.46}$ & $83.61 \pm 0.78$ & $86.88 \pm 0.67$ \\
Full: no spectral normalization & $58.04 \pm 1.02$ & $76.53 \pm 0.44$ & $81.50 \pm 0.11$ & $56.27 \pm 2.10$ & $77.74 \pm 0.34$ & $82.29 \pm 0.75$ \\
\midrule
\multicolumn{7}{l}{\textit{Reliability-gate ablations}} \\
Full: disagreement only & $\mathbf{62.22 \pm 1.11}$ & $\underline{83.53 \pm 0.81}$ & $\mathbf{85.71 \pm 0.12}$ & $\mathbf{62.12 \pm 1.43}$ & $83.79 \pm 0.79$ & $\mathbf{87.09 \pm 0.64}$ \\
Full: entropy only & $\underline{62.04 \pm 1.07}$ & $\mathbf{83.59 \pm 0.80}$ & $\underline{85.70 \pm 0.22}$ & $61.80 \pm 1.46$ & $\mathbf{83.85 \pm 0.83}$ & $\underline{87.02 \pm 0.50}$ \\
\midrule
\multicolumn{7}{l}{\textit{Complete framework}} \\
Full BIDETA & $\mathbf{62.22 \pm 1.11}$ & $\mathbf{83.59 \pm 0.81}$ & $\mathbf{85.71 \pm 0.12}$ & $\mathbf{62.12 \pm 1.43}$ & $\underline{83.81 \pm 0.80}$ & $\mathbf{87.09 \pm 0.61}$ \\
\bottomrule
\end{tabular}
\end{adjustbox}
\end{table}

\subsection{Closed-set identity ranking}
Table~\ref{tab:tacquad} reports closed-set identity-ranking performance on TacQuad with different pretrained backbones. Using the TVL backbone, BIDETA improves MRR over the frozen backbone by an average of 64.25 percentage points and over the strongest comparison by an average of 8.45 percentage points. Using Sparsh, BIDETA outperforms the strongest external comparison by 7.55, 12.89, and 14.86 percentage points at the 10\%, 20\%, and 30\% support budgets, respectively. These results demonstrate BIDETA's cross-task gains, validating the framework's adaptability beyond classification. Appendix~\ref{app:ranking} reports additional R@1 results.

\subsection{Ablation Study}

Table~\ref{tab:ablation} reports the component ablations on SITR. Among the sequential component additions, consensus recurrence produces the largest gain, improving accuracy by 2.65--4.71 percentage points across the six backbone--budget settings. By repeatedly aggregating cross-view-consistent evidence over the query graph, it strengthens neighborhood coherence and class discrimination. Within the full model, removing spectral normalization causes the largest performance loss, reducing accuracy by 5.36 percentage points on average and by up to 7.06 points. Its support-conditioned covariance transformation suppresses sensor-dependent variation and aligns target neighborhoods with class structure, making it central to robust cross-sensor propagation. The view-1-only and view-2-only configurations also show a clear decline relative to full BIDETA, indicating that complementary hierarchical representations preserve richer class evidence and stabilize data-efficient adaptation. Together, these components provide complementary functions, and their integration achieves the strongest overall performance across the evaluated settings. Appendix~\ref{app:ablation} reports additional macro-F1 ablations, while Appendix~\ref{app:sensitivity} presents the parameter-sensitivity experiments.
\FloatBarrier
 
\section{Conclusion}
Tactile foundation models suffer substantial performance degradation when deployed with sensors that differ from those seen during pretraining. To address this problem, we proposed BIDETA, a brain-inspired data-efficient adaptation framework that adapts frozen tactile encoders to unseen sensors through rapid support memory, support-conditioned spectral geometry, and reliability-gated recurrence. Experiments across multiple benchmarks show that BIDETA effectively improves the performance of tactile foundation models on unseen sensors and transfers these gains across datasets, foundation models, and tasks.
Future work will extend BIDETA to online robotic manipulation, visual--tactile material understanding, and distributed tactile-skin perception. These advances can promote sensor-agnostic tactile foundation models and address the rapid transfer of embodied agents to newly deployed, heterogeneous tactile hardware.

\label{maintext:end}
\bibliography{references}
\bibliographystyle{plainnat}
\clearpage
\appendix
\section{Protocol and implementation details}
\label{app:protocol}
\subsection{Algorithm Procedure}
\label{app:algorithm}
\begin{algorithm}[!htbp]
\caption{BIDETA adaptation for one target sensor}
\label{alg:BIDETA}
\begin{algorithmic}[1]
\Require Frozen encoder $f_\theta$, labeled support $\mathcal{S}$, unlabeled queries $\mathcal{Q}$, and fixed hyperparameters
\State Extract two support and query readouts with $f_\theta$
\For{$v\in\{1,2\}$}
 \State Fit the support scaler and shrinkage-LDA memory
 \State Compute query probabilities $P^{(v)}$ and spectral transform $A^{(v)}$
 \State Transform query features and construct the readout graph $G^{(v)}$
\EndFor
\State Fuse the two memories into $P_0$ and the two graphs into $G$
\State Compute the reliability gate $R$ and initialize $P\gets P_0$
\For{$t=1,\ldots,T$}
 \State $P\gets\NormPrior[(I_m-R)P_0+RGP]$
\EndFor
\State \Return class predictions or identity rankings from $P$
\end{algorithmic}
\end{algorithm}
\FloatBarrier

\subsection{Dataset splits and sample accounting}
\label{app:data} 
SITR's original classification archive contains 20 classes across seven sensors. We follow the supplied 16-class subset, mapped consecutively to the classifier label space. This yields 112,000 images: 12,800 training and 3,200 validation images per sensor. The six target episodes each use 128, 640, or 1,280 support images. The source is DIGIT, and targets correspond to Mini\_1, Mini\_2, Mini\_3, Mini\_4, Hex, and Wedge. The train-only inner split is distinct from this evaluation set; the final support sample is drawn from the full 800-per-class training pool.

The TacVerse Shape archive contains 30,094 images across seven sensors. The experiment uses the source plus five targets specified in the main text; GelSightMarker is not part of the five-target average. The nine labels are column, cuboid, dots, edge, hexagon, moon, ring, triangles, and wave. For the included sensor--class pairs, the 300/100/100 ordered split yields 2,700 training images, 900 validation images, and 900 test images per sensor. The support sizes are 27, 135, and 270 per target.

The final TacQuad task retains 56 trial identities with complete 20-frame sequences across all three RGB sensors. The force-field sensor Tac3D is excluded from the RGB encoder experiment. The source contains 12 frames per identity (672 total); target support contains 112/224/336 frames, and each target query set contains 224 frames. The support and query positions are disjoint within the sampled sequence, but they share identity and can share a physical trial. Table~\ref{tab:datasetlabels} summarizes the label spaces used by all three benchmarks.

\subsection{Preprocessing and Feature Extraction}
\label{app:preprocessing}
SITR and TacVerse images are converted to RGB and resized to $224\times224$ with PIL bicubic interpolation. TVL uses channel means $(0.291746,0.297133,0.291040)$ and standard deviations $(0.187645,0.194677,0.218716)$ after conversion to $[0,1]$. Sparsh uses no additional channel normalization and concatenates the image with itself to obtain six channels. TacQuad uses torchvision resizing with bilinear interpolation and antialiasing.

The TVL wrapper loads the tactile-encoder state from the released checkpoint and returns global average-pooled and projected features. The first readout is not a penultimate transformer block or a CLS token. For Sparsh, the wrapper mean-pools normalized patch tokens from the final two blocks, excluding register tokens. The archived frozen parameter counts are 21,961,344 for TVL and 21,894,144 for Sparsh. Encoders run in evaluation mode without gradient updates.

\begin{table}[!htbp]
\centering
\caption{\textbf{Label spaces used by the three benchmarks.} The table summarizes the classification labels and enrolled identity definition used in our experiments.}
\label{tab:datasetlabels}
\setlength{\tabcolsep}{4pt}
\renewcommand{\arraystretch}{1.08}
\begin{tabular}{p{0.14\linewidth}p{0.78\linewidth}}
\toprule
Dataset & Classes or identities used \\
\midrule
SITR & 16 household-object classes selected from the 20-object classification release. \\
TacVerse Shape & Nine local-shape classes: column, cuboid, dots, edge, hexagon, moon, ring, triangles, and wave. \\
TacQuad & 56 trial-folder identities. Each identity denotes one enrolled object-contact trial represented by a 20-frame sequence for each RGB sensor. \\
\bottomrule
\end{tabular}
\end{table}

\begin{table}[!htbp]
\centering
\caption{Baseline implementations.}
\label{tab:permissions}
\begin{adjustbox}{max width=\linewidth}
\begin{tabular}{llll}
\toprule
Method & Readout & Target fitting & Joint query inference\\
\midrule
Frozen backbone & Final & None & No\\
Tip-Adapter & Final & Support cache & No\\
SimpleShot & Earlier & Class prototypes & No\\
LaplacianShot & Earlier & Class prototypes & Yes\\
SITR-Calib & Final & Lightweight head & No\\
AnyTouch Match & Final & Lightweight head & No\\
CTSRL CSM & Final & Lightweight head & No\\
BIDETA & Both & Statistical memory and graph & Yes\\
\bottomrule
\end{tabular}
\end{adjustbox}
\end{table}

\begin{table}[!htbp]
\centering
\caption{\textbf{Hyperparameter configurations used in the experiments.} SITR and TacVerse share one configuration, while TacQuad uses a task-specific configuration.}
\label{tab:hyperparameters}
\small
\begin{tabular}{lc}
\toprule
Hyperparameter & Value \\
\midrule
\multicolumn{2}{l}{\textbf{SITR and TacVerse}} \\
Spectral exponent ($\gamma$) & 0.6 \\
Neighborhood size ($k$) & 40 \\
Graph temperature ($\tau_g$) & 0.2 \\
Recurrence range ($r_{\min}$--$r_{\max}$) & 0.70--0.90 \\
Iterations ($T$) & 10 \\
Disagreement weight ($\lambda$) & 1.0 \\
Gate exponent ($p$) & 0.05 \\
\midrule
\multicolumn{2}{l}{\textbf{TacQuad}} \\
Spectral exponent ($\gamma$) & 0.25 \\
Neighborhood size ($k$) & 80 \\
Graph temperature ($\tau_g$) & 0.07 \\
Recurrence range ($r_{\min}$--$r_{\max}$) & 0.60--0.80 \\
Iterations ($T$) & 20 \\
Disagreement weight ($\lambda$) & 0.50 \\
Gate exponent ($p$) & 0.15 \\
\bottomrule
\end{tabular}
\end{table}

\subsection{Baseline Implementations}
\label{app:baselines}

Methods with lightweight classification heads keep the tactile encoder frozen and optimize compact feature modules and classifiers with AdamW~\citep{loshchilov2019adamw}, class-balanced target sampling, target cross-entropy, source-prototype preservation, and within-class compactness. The shared settings use a weight decay of 0.0005, at most 300 epochs, and 16 sampled examples per class in each training step. Checkpoints are selected using the training loss. Frozen-feature methods use normalized source or support features, class prototypes, support caches, or query graphs without updating the backbone.

\subsection{Experimental Settings and Hyperparameters}
\label{app:settings}
SITR and TacVerse use the same hyperparameter configuration, while TacQuad uses a separate configuration because it evaluates identity ranking. Table~\ref{tab:hyperparameters} lists the fixed values used for these two settings.

\section{Classification Results}
\label{app:classification}
\subsection{Per-Sensor Accuracy Results}
Tables~\ref{tab:detail1}--\ref{tab:detail4} report all main accuracy results by sensor. The gains generally increase with the support budget, and BIDETA achieves the strongest overall performance across sensors and backbones, showing that support-conditioned memory and graph inference effectively adapt pretrained representations to sensor-specific shifts.
\begin{table}[!htbp]
\centering
\caption{\textbf{Per-sensor accuracy on SITR with TVL.} Results (\%) are reported as mean $\pm$ sample standard deviation over three random seeds. Best results are in bold and second-best results are underlined.}
\label{tab:detail1}
\scriptsize
\setlength{\tabcolsep}{3pt}
\begin{adjustbox}{max width=\linewidth}
\begin{tabular}{lcccccccc}
\toprule
Method & Labels & Mini 1 & Mini 2 & Mini 3 & Mini 4 & Hex & Wedge & Mean \\
\midrule
Frozen backbone & 0\% & $9.27 \pm 0.81$ & $7.89 \pm 0.13$ & $6.62 \pm 0.16$ & $6.10 \pm 0.05$ & $6.22 \pm 0.05$ & $9.39 \pm 0.25$ & $7.58 \pm 0.14$ \\
Tip-Adapter & 1\% & $12.40 \pm 5.43$ & $24.88 \pm 6.23$ & $37.03 \pm 1.09$ & $13.27 \pm 0.21$ & $6.22 \pm 0.05$ & $9.48 \pm 0.34$ & $17.21 \pm 1.63$ \\
Tip-Adapter & 5\% & $23.33 \pm 0.24$ & $31.39 \pm 1.26$ & $45.15 \pm 0.21$ & $23.54 \pm 2.32$ & $6.25 \pm 0.00$ & $11.43 \pm 3.16$ & $23.51 \pm 0.51$ \\
Tip-Adapter & 10\% & $25.74 \pm 2.32$ & $34.50 \pm 0.68$ & $44.39 \pm 1.87$ & $28.26 \pm 0.82$ & $7.54 \pm 1.12$ & $30.96 \pm 4.38$ & $28.56 \pm 1.37$ \\
SimpleShot & 1\% & $33.14 \pm 3.93$ & $32.93 \pm 3.49$ & $50.80 \pm 1.01$ & $34.61 \pm 5.27$ & $34.28 \pm 2.15$ & $\underline{42.15 \pm 0.85}$ & $37.98 \pm 0.69$ \\
SimpleShot & 5\% & $41.25 \pm 2.24$ & $41.29 \pm 1.32$ & $58.63 \pm 1.52$ & $41.00 \pm 1.81$ & $37.88 \pm 0.68$ & $46.75 \pm 1.26$ & $44.47 \pm 0.90$ \\
SimpleShot & 10\% & $39.81 \pm 0.89$ & $43.28 \pm 2.18$ & $58.11 \pm 2.02$ & $40.65 \pm 1.17$ & $\underline{37.27 \pm 0.95}$ & $\underline{48.19 \pm 0.46}$ & $44.55 \pm 0.62$ \\
LaplacianShot & 1\% & $29.73 \pm 3.37$ & $36.61 \pm 4.43$ & $55.74 \pm 1.39$ & $32.80 \pm 5.88$ & $\underline{35.30 \pm 2.63}$ & $41.08 \pm 1.12$ & $\underline{38.55 \pm 0.38}$ \\
LaplacianShot & 5\% & $42.02 \pm 3.48$ & $44.86 \pm 0.83$ & $61.02 \pm 2.57$ & $40.72 \pm 1.37$ & $\underline{38.05 \pm 1.47}$ & $\underline{48.36 \pm 1.55}$ & $45.84 \pm 1.65$ \\
LaplacianShot & 10\% & $37.23 \pm 2.03$ & $44.25 \pm 3.34$ & $60.97 \pm 3.64$ & $38.07 \pm 0.83$ & $37.11 \pm 2.06$ & $47.02 \pm 1.50$ & $44.11 \pm 0.60$ \\
SITR-Calib & 1\% & $40.03 \pm 1.89$ & $\underline{36.83 \pm 2.08}$ & $\underline{55.78 \pm 1.54}$ & $\underline{39.97 \pm 4.84}$ & $17.81 \pm 0.20$ & $37.56 \pm 3.98$ & $38.00 \pm 1.56$ \\
SITR-Calib & 5\% & $46.85 \pm 3.45$ & $56.40 \pm 0.97$ & $68.15 \pm 4.64$ & $41.46 \pm 1.85$ & $23.66 \pm 3.72$ & $42.95 \pm 1.47$ & $\underline{46.58 \pm 0.08}$ \\
SITR-Calib & 10\% & $\underline{46.10 \pm 1.75}$ & $\underline{64.46 \pm 1.31}$ & $68.43 \pm 2.26$ & $41.97 \pm 1.88$ & $21.19 \pm 6.03$ & $42.65 \pm 2.22$ & $\underline{47.47 \pm 1.44}$ \\
AnyTouch Match & 1\% & $\underline{41.98 \pm 1.81}$ & $35.20 \pm 2.10$ & $55.30 \pm 1.44$ & $38.81 \pm 5.66$ & $6.03 \pm 1.22$ & $40.27 \pm 2.70$ & $36.27 \pm 1.55$ \\
AnyTouch Match & 5\% & $\underline{47.41 \pm 2.19}$ & $56.49 \pm 0.66$ & $\underline{70.41 \pm 2.61}$ & $\underline{42.44 \pm 1.49}$ & $6.58 \pm 0.73$ & $42.50 \pm 0.78$ & $44.30 \pm 0.12$ \\
AnyTouch Match & 10\% & $44.07 \pm 3.38$ & $63.97 \pm 1.00$ & $\underline{68.68 \pm 1.64}$ & $\underline{44.22 \pm 1.90}$ & $6.07 \pm 1.71$ & $44.90 \pm 2.89$ & $45.32 \pm 1.00$ \\
CTSRL CSM & 1\% & $39.93 \pm 0.89$ & $36.71 \pm 2.04$ & $55.59 \pm 2.86$ & $36.60 \pm 5.43$ & $17.61 \pm 3.08$ & $38.09 \pm 0.23$ & $37.42 \pm 1.77$ \\
CTSRL CSM & 5\% & $44.99 \pm 1.42$ & $\underline{56.72 \pm 1.11}$ & $66.64 \pm 3.83$ & $41.82 \pm 2.05$ & $15.79 \pm 2.71$ & $37.28 \pm 1.41$ & $43.87 \pm 1.29$ \\
CTSRL CSM & 10\% & $45.24 \pm 1.83$ & $63.53 \pm 1.02$ & $65.88 \pm 2.70$ & $41.40 \pm 1.75$ & $15.72 \pm 4.68$ & $38.21 \pm 3.24$ & $44.99 \pm 0.85$ \\
BIDETA & 1\% & $\mathbf{53.39 \pm 3.20}$ & $\mathbf{52.66 \pm 0.65}$ & $\mathbf{72.06 \pm 3.99}$ & $\mathbf{62.18 \pm 1.84}$ & $\mathbf{65.55 \pm 4.27}$ & $\mathbf{67.47 \pm 2.32}$ & $\mathbf{62.22 \pm 1.11}$ \\
BIDETA & 5\% & $\mathbf{78.33 \pm 1.77}$ & $\mathbf{78.88 \pm 1.06}$ & $\mathbf{88.05 \pm 2.62}$ & $\mathbf{81.63 \pm 1.21}$ & $\mathbf{82.72 \pm 1.89}$ & $\mathbf{91.93 \pm 1.58}$ & $\mathbf{83.59 \pm 0.81}$ \\
BIDETA & 10\% & $\mathbf{81.13 \pm 0.83}$ & $\mathbf{84.41 \pm 2.67}$ & $\mathbf{88.25 \pm 0.94}$ & $\mathbf{79.85 \pm 1.99}$ & $\mathbf{86.50 \pm 3.35}$ & $\mathbf{94.10 \pm 0.83}$ & $\mathbf{85.71 \pm 0.12}$ \\
\bottomrule
\end{tabular}
\end{adjustbox}
\end{table}

\begin{table}[!htbp]
\centering
\caption{\textbf{Per-sensor accuracy on SITR with Sparsh.} Results (\%) are reported as mean $\pm$ sample standard deviation over three random seeds. Best results are in bold and second-best results are underlined.}
\label{tab:detail2}
\scriptsize
\setlength{\tabcolsep}{3pt}
\begin{adjustbox}{max width=\linewidth}
\begin{tabular}{lcccccccc}
\toprule
Method & Labels & Mini 1 & Mini 2 & Mini 3 & Mini 4 & Hex & Wedge & Mean \\
\midrule
Frozen backbone & 0\% & $6.25 \pm 0.00$ & $6.25 \pm 0.00$ & $6.25 \pm 0.00$ & $6.25 \pm 0.00$ & $6.26 \pm 0.02$ & $9.88 \pm 1.99$ & $6.86 \pm 0.33$ \\
Tip-Adapter & 1\% & $6.86 \pm 1.06$ & $27.56 \pm 1.70$ & $32.04 \pm 2.51$ & $6.25 \pm 0.00$ & $6.26 \pm 0.02$ & $10.13 \pm 1.96$ & $14.85 \pm 0.34$ \\
Tip-Adapter & 5\% & $20.83 \pm 0.46$ & $30.58 \pm 2.86$ & $38.05 \pm 1.81$ & $23.00 \pm 2.89$ & $6.27 \pm 0.04$ & $12.67 \pm 0.69$ & $21.90 \pm 0.92$ \\
Tip-Adapter & 10\% & $20.40 \pm 0.93$ & $34.23 \pm 1.76$ & $36.54 \pm 1.19$ & $22.44 \pm 0.74$ & $6.58 \pm 0.58$ & $13.59 \pm 0.57$ & $22.30 \pm 0.04$ \\
SimpleShot & 1\% & $25.53 \pm 2.51$ & $32.70 \pm 3.68$ & $37.35 \pm 4.25$ & $28.97 \pm 5.44$ & $\underline{29.85 \pm 2.83}$ & $\underline{38.07 \pm 1.88}$ & $32.08 \pm 2.45$ \\
SimpleShot & 5\% & $33.40 \pm 2.23$ & $36.18 \pm 2.78$ & $46.02 \pm 1.49$ & $35.98 \pm 2.85$ & $\underline{34.81 \pm 4.33}$ & $45.46 \pm 0.30$ & $38.64 \pm 0.96$ \\
SimpleShot & 10\% & $31.57 \pm 1.07$ & $41.28 \pm 3.01$ & $43.52 \pm 0.13$ & $35.80 \pm 1.72$ & $\underline{37.23 \pm 0.51}$ & $\underline{45.84 \pm 0.66}$ & $39.21 \pm 0.81$ \\
LaplacianShot & 1\% & $25.30 \pm 3.18$ & $35.99 \pm 5.79$ & $39.10 \pm 6.45$ & $29.20 \pm 6.45$ & $29.82 \pm 5.30$ & $37.26 \pm 2.77$ & $\underline{32.78 \pm 3.52}$ \\
LaplacianShot & 5\% & $32.82 \pm 2.79$ & $40.30 \pm 2.88$ & $47.49 \pm 1.85$ & $\underline{38.13 \pm 4.49}$ & $34.68 \pm 4.93$ & $\underline{46.20 \pm 0.69}$ & $\underline{39.94 \pm 1.29}$ \\
LaplacianShot & 10\% & $31.72 \pm 1.14$ & $\underline{44.97 \pm 4.40}$ & $43.71 \pm 1.37$ & $37.30 \pm 3.74$ & $36.08 \pm 0.48$ & $45.43 \pm 0.34$ & $\underline{39.87 \pm 0.77}$ \\
SITR-Calib & 1\% & $33.19 \pm 5.49$ & $35.42 \pm 1.87$ & $44.82 \pm 4.49$ & $28.85 \pm 1.40$ & $13.96 \pm 1.66$ & $29.01 \pm 2.12$ & $30.88 \pm 1.99$ \\
SITR-Calib & 5\% & $36.48 \pm 0.10$ & $36.52 \pm 3.99$ & $52.08 \pm 6.81$ & $35.95 \pm 0.81$ & $14.70 \pm 2.21$ & $31.02 \pm 0.44$ & $34.46 \pm 1.00$ \\
SITR-Calib & 10\% & $36.07 \pm 1.39$ & $39.01 \pm 1.25$ & $50.68 \pm 2.88$ & $36.67 \pm 2.00$ & $13.41 \pm 0.61$ & $29.17 \pm 2.82$ & $34.17 \pm 0.35$ \\
AnyTouch Match & 1\% & $\underline{38.15 \pm 4.88}$ & $\underline{36.82 \pm 2.23}$ & $\underline{45.34 \pm 4.25}$ & $\underline{32.96 \pm 0.97}$ & $11.66 \pm 2.00$ & $29.09 \pm 2.72$ & $32.34 \pm 0.99$ \\
AnyTouch Match & 5\% & $\underline{40.17 \pm 1.30}$ & $\underline{41.64 \pm 2.01}$ & $52.00 \pm 0.97$ & $37.58 \pm 1.82$ & $12.28 \pm 1.26$ & $28.51 \pm 4.11$ & $35.36 \pm 0.56$ \\
AnyTouch Match & 10\% & $\underline{38.60 \pm 2.37}$ & $43.94 \pm 2.02$ & $\underline{54.44 \pm 4.56}$ & $\underline{38.32 \pm 0.35}$ & $12.94 \pm 1.73$ & $26.63 \pm 4.75$ & $35.81 \pm 0.64$ \\
CTSRL CSM & 1\% & $34.83 \pm 1.75$ & $34.59 \pm 2.57$ & $44.85 \pm 3.85$ & $30.23 \pm 1.27$ & $16.40 \pm 1.02$ & $30.96 \pm 1.13$ & $31.98 \pm 1.27$ \\
CTSRL CSM & 5\% & $37.95 \pm 0.23$ & $38.85 \pm 1.38$ & $\underline{54.39 \pm 2.85}$ & $34.11 \pm 1.63$ & $14.22 \pm 1.30$ & $30.10 \pm 1.23$ & $34.94 \pm 0.25$ \\
CTSRL CSM & 10\% & $36.16 \pm 0.69$ & $38.51 \pm 1.95$ & $51.66 \pm 2.16$ & $35.16 \pm 1.34$ & $13.70 \pm 1.94$ & $30.70 \pm 1.81$ & $34.31 \pm 0.39$ \\
BIDETA & 1\% & $\mathbf{51.34 \pm 2.86}$ & $\mathbf{63.73 \pm 3.73}$ & $\mathbf{63.06 \pm 4.50}$ & $\mathbf{55.88 \pm 3.19}$ & $\mathbf{65.64 \pm 4.47}$ & $\mathbf{73.05 \pm 2.49}$ & $\mathbf{62.12 \pm 1.43}$ \\
BIDETA & 5\% & $\mathbf{67.35 \pm 2.36}$ & $\mathbf{93.66 \pm 1.13}$ & $\mathbf{80.07 \pm 2.04}$ & $\mathbf{82.10 \pm 1.13}$ & $\mathbf{89.10 \pm 1.17}$ & $\mathbf{90.59 \pm 1.69}$ & $\mathbf{83.81 \pm 0.80}$ \\
BIDETA & 10\% & $\mathbf{71.52 \pm 2.61}$ & $\mathbf{95.89 \pm 0.10}$ & $\mathbf{84.66 \pm 0.81}$ & $\mathbf{86.21 \pm 1.11}$ & $\mathbf{92.88 \pm 0.81}$ & $\mathbf{91.42 \pm 1.17}$ & $\mathbf{87.09 \pm 0.61}$ \\
\bottomrule
\end{tabular}
\end{adjustbox}
\end{table}

\begin{table}[!htbp]
\centering
\caption{\textbf{Per-sensor accuracy on TacVerse Shape with TVL.} Results (\%) are reported as mean $\pm$ sample standard deviation over three random seeds. Best results are in bold and second-best results are underlined. SITR-Calib is unavailable because standard calibration images are absent.}
\label{tab:detail3}
\scriptsize
\setlength{\tabcolsep}{3pt}
\begin{adjustbox}{max width=\linewidth}
\begin{tabular}{lccccccc}
\toprule
Method & Labels & MagicGripper & MagicTac & TacTip & ViTac & ViTacTip & Mean \\
\midrule
Frozen backbone & 0\% & $8.67 \pm 0.38$ & $6.48 \pm 1.80$ & $10.41 \pm 1.00$ & $11.11 \pm 0.00$ & $11.11 \pm 0.00$ & $9.56 \pm 0.48$ \\
Tip-Adapter & 1\% & $23.07 \pm 6.55$ & $30.26 \pm 13.45$ & $10.74 \pm 1.12$ & $26.56 \pm 14.11$ & $23.30 \pm 2.26$ & $22.79 \pm 4.24$ \\
Tip-Adapter & 5\% & $39.78 \pm 3.84$ & $52.07 \pm 0.84$ & $17.48 \pm 9.36$ & $43.37 \pm 1.99$ & $44.56 \pm 0.69$ & $39.45 \pm 2.15$ \\
Tip-Adapter & 10\% & $42.41 \pm 3.81$ & $51.19 \pm 0.53$ & $28.93 \pm 0.84$ & $43.81 \pm 1.53$ & $61.26 \pm 1.57$ & $45.52 \pm 0.73$ \\
SimpleShot & 1\% & $33.78 \pm 5.61$ & $46.59 \pm 4.33$ & $\underline{29.56 \pm 2.19}$ & $\underline{55.00 \pm 6.47}$ & $\underline{57.93 \pm 7.47}$ & $\underline{44.57 \pm 2.70}$ \\
SimpleShot & 5\% & $40.78 \pm 3.72$ & $47.67 \pm 0.19$ & $\underline{33.78 \pm 1.64}$ & $64.00 \pm 2.98$ & $\underline{69.41 \pm 3.49}$ & $51.13 \pm 1.73$ \\
SimpleShot & 10\% & $48.85 \pm 3.04$ & $48.30 \pm 1.67$ & $\underline{36.11 \pm 0.40}$ & $64.22 \pm 3.61$ & $69.26 \pm 2.64$ & $53.35 \pm 0.76$ \\
LaplacianShot & 1\% & $\underline{36.56 \pm 8.76}$ & $47.52 \pm 2.31$ & $26.22 \pm 5.46$ & $52.15 \pm 6.20$ & $55.41 \pm 4.01$ & $43.57 \pm 4.28$ \\
LaplacianShot & 5\% & $41.85 \pm 6.41$ & $46.37 \pm 1.68$ & $31.07 \pm 4.59$ & $\underline{66.30 \pm 4.07}$ & $62.70 \pm 3.95$ & $49.66 \pm 1.61$ \\
LaplacianShot & 10\% & $49.89 \pm 3.01$ & $47.67 \pm 2.38$ & $34.00 \pm 0.56$ & $65.04 \pm 2.81$ & $61.44 \pm 4.46$ & $51.61 \pm 1.26$ \\
SITR-Calib & 1\% & -- & -- & -- & -- & -- & -- \\
SITR-Calib & 5\% & -- & -- & -- & -- & -- & -- \\
SITR-Calib & 10\% & -- & -- & -- & -- & -- & -- \\
AnyTouch Match & 1\% & $29.67 \pm 4.45$ & $54.44 \pm 11.26$ & $20.59 \pm 1.61$ & $52.78 \pm 6.44$ & $54.70 \pm 8.56$ & $42.44 \pm 1.93$ \\
AnyTouch Match & 5\% & $41.37 \pm 8.66$ & $70.15 \pm 3.02$ & $22.41 \pm 0.96$ & $60.26 \pm 4.72$ & $63.85 \pm 4.72$ & $51.61 \pm 1.73$ \\
AnyTouch Match & 10\% & $\underline{50.89 \pm 1.49}$ & $70.74 \pm 2.84$ & $22.48 \pm 1.03$ & $\underline{68.04 \pm 0.23}$ & $67.00 \pm 2.67$ & $55.83 \pm 0.37$ \\
CTSRL CSM & 1\% & $30.33 \pm 2.99$ & $\underline{54.93 \pm 14.75}$ & $22.56 \pm 3.53$ & $54.63 \pm 5.14$ & $53.85 \pm 8.46$ & $43.26 \pm 1.41$ \\
CTSRL CSM & 5\% & $\underline{42.30 \pm 8.79}$ & $\underline{72.70 \pm 2.56}$ & $25.19 \pm 1.67$ & $62.19 \pm 0.93$ & $65.67 \pm 6.01$ & $\underline{53.61 \pm 3.41}$ \\
CTSRL CSM & 10\% & $50.22 \pm 2.83$ & $\underline{71.74 \pm 0.68}$ & $25.33 \pm 1.07$ & $65.26 \pm 0.97$ & $\underline{70.78 \pm 0.69}$ & $\underline{56.67 \pm 0.83}$ \\
BIDETA & 1\% & $\mathbf{38.41 \pm 6.56}$ & $\mathbf{73.59 \pm 7.58}$ & $\mathbf{36.93 \pm 2.21}$ & $\mathbf{73.30 \pm 3.22}$ & $\mathbf{67.44 \pm 3.70}$ & $\mathbf{57.93 \pm 1.73}$ \\
BIDETA & 5\% & $\mathbf{52.44 \pm 5.17}$ & $\mathbf{99.44 \pm 0.87}$ & $\mathbf{42.74 \pm 1.39}$ & $\mathbf{95.85 \pm 0.65}$ & $\mathbf{86.89 \pm 4.26}$ & $\mathbf{75.47 \pm 2.03}$ \\
BIDETA & 10\% & $\mathbf{57.04 \pm 6.40}$ & $\mathbf{99.48 \pm 0.71}$ & $\mathbf{51.52 \pm 3.21}$ & $\mathbf{97.52 \pm 1.07}$ & $\mathbf{90.11 \pm 1.16}$ & $\mathbf{79.13 \pm 1.20}$ \\
\bottomrule
\end{tabular}
\end{adjustbox}
\end{table}

\begin{table}[!htbp]
\centering
\caption{\textbf{Per-sensor accuracy on TacVerse Shape with Sparsh.} Results (\%) are reported as mean $\pm$ sample standard deviation over three random seeds. Best results are in bold and second-best results are underlined. SITR-Calib is unavailable because standard calibration images are absent.}
\label{tab:detail4}
\scriptsize
\setlength{\tabcolsep}{3pt}
\begin{adjustbox}{max width=\linewidth}
\begin{tabular}{lccccccc}
\toprule
Method & Labels & MagicGripper & MagicTac & TacTip & ViTac & ViTacTip & Mean \\
\midrule
Frozen backbone & 0\% & $10.59 \pm 0.63$ & $11.11 \pm 0.00$ & $11.11 \pm 0.00$ & $17.26 \pm 1.58$ & $18.26 \pm 1.98$ & $13.67 \pm 0.47$ \\
Tip-Adapter & 1\% & $13.19 \pm 3.39$ & $11.11 \pm 0.00$ & $11.11 \pm 0.00$ & $17.48 \pm 1.45$ & $18.33 \pm 2.04$ & $14.24 \pm 1.07$ \\
Tip-Adapter & 5\% & $43.70 \pm 2.84$ & $11.11 \pm 0.00$ & $11.11 \pm 0.00$ & $31.15 \pm 0.93$ & $30.96 \pm 0.50$ & $25.61 \pm 0.74$ \\
Tip-Adapter & 10\% & $43.93 \pm 1.62$ & $57.33 \pm 2.44$ & $11.11 \pm 0.00$ & $32.48 \pm 0.51$ & $32.30 \pm 0.45$ & $35.43 \pm 0.72$ \\
SimpleShot & 1\% & $38.56 \pm 2.11$ & $\underline{48.04 \pm 4.45}$ & $\mathbf{40.00 \pm 5.14}$ & $57.96 \pm 6.42$ & $48.48 \pm 1.01$ & $\underline{46.61 \pm 0.54}$ \\
SimpleShot & 5\% & $48.70 \pm 2.41$ & $58.15 \pm 2.06$ & $41.00 \pm 0.87$ & $\underline{68.48 \pm 1.10}$ & $52.67 \pm 1.95$ & $\underline{53.80 \pm 0.91}$ \\
SimpleShot & 10\% & $48.85 \pm 2.39$ & $\underline{59.70 \pm 2.45}$ & $\underline{46.78 \pm 2.35}$ & $\underline{69.93 \pm 1.56}$ & $54.63 \pm 1.17$ & $\underline{55.98 \pm 0.86}$ \\
LaplacianShot & 1\% & $38.41 \pm 2.93$ & $46.74 \pm 5.47$ & $38.93 \pm 4.14$ & $56.22 \pm 8.11$ & $44.52 \pm 4.39$ & $44.96 \pm 2.31$ \\
LaplacianShot & 5\% & $48.07 \pm 3.03$ & $\underline{58.37 \pm 2.17}$ & $\underline{41.44 \pm 2.70}$ & $64.22 \pm 0.69$ & $45.26 \pm 5.20$ & $51.47 \pm 0.60$ \\
LaplacianShot & 10\% & $47.56 \pm 2.22$ & $58.00 \pm 3.07$ & $44.48 \pm 2.97$ & $68.81 \pm 3.11$ & $42.37 \pm 2.45$ & $52.24 \pm 1.17$ \\
SITR-Calib & 1\% & -- & -- & -- & -- & -- & -- \\
SITR-Calib & 5\% & -- & -- & -- & -- & -- & -- \\
SITR-Calib & 10\% & -- & -- & -- & -- & -- & -- \\
AnyTouch Match & 1\% & $\underline{41.15 \pm 2.44}$ & $45.63 \pm 6.30$ & $11.11 \pm 0.00$ & $58.22 \pm 3.66$ & $50.52 \pm 1.69$ & $41.33 \pm 1.71$ \\
AnyTouch Match & 5\% & $\underline{58.00 \pm 3.29}$ & $54.52 \pm 1.12$ & $11.11 \pm 0.00$ & $64.37 \pm 1.83$ & $\underline{58.15 \pm 2.02}$ & $49.23 \pm 1.16$ \\
AnyTouch Match & 10\% & $\underline{58.74 \pm 3.11}$ & $57.70 \pm 3.24$ & $11.11 \pm 0.00$ & $66.11 \pm 4.03$ & $55.44 \pm 4.35$ & $49.82 \pm 2.12$ \\
CTSRL CSM & 1\% & $40.67 \pm 6.30$ & $44.74 \pm 6.53$ & $12.96 \pm 0.65$ & $\mathbf{61.15 \pm 3.73}$ & $\underline{52.67 \pm 1.95}$ & $42.44 \pm 1.93$ \\
CTSRL CSM & 5\% & $53.96 \pm 5.11$ & $52.89 \pm 1.15$ & $14.81 \pm 3.61$ & $68.04 \pm 1.34$ & $57.81 \pm 1.39$ & $49.50 \pm 1.82$ \\
CTSRL CSM & 10\% & $56.63 \pm 3.11$ & $52.67 \pm 3.42$ & $13.59 \pm 0.93$ & $68.19 \pm 0.34$ & $\underline{57.63 \pm 3.34}$ & $49.74 \pm 1.97$ \\
BIDETA & 1\% & $\mathbf{49.22 \pm 8.19}$ & $\mathbf{89.37 \pm 5.11}$ & $\underline{39.00 \pm 2.44}$ & $\underline{58.74 \pm 2.98}$ & $\mathbf{61.96 \pm 6.10}$ & $\mathbf{59.66 \pm 2.61}$ \\
BIDETA & 5\% & $\mathbf{81.78 \pm 7.62}$ & $\mathbf{100.00 \pm 0.00}$ & $\mathbf{52.30 \pm 1.88}$ & $\mathbf{88.26 \pm 1.98}$ & $\mathbf{80.93 \pm 3.37}$ & $\mathbf{80.65 \pm 0.43}$ \\
BIDETA & 10\% & $\mathbf{89.74 \pm 3.26}$ & $\mathbf{100.00 \pm 0.00}$ & $\mathbf{62.48 \pm 1.80}$ & $\mathbf{87.63 \pm 1.62}$ & $\mathbf{78.04 \pm 5.65}$ & $\mathbf{83.58 \pm 1.05}$ \\
\bottomrule
\end{tabular}
\end{adjustbox}
\end{table}

\subsection{Macro-F1 and sensor dependence}
Macro-F1 assigns equal weight to every class and complements accuracy. Tables~\ref{tab:detail1f1}--\ref{tab:detail4f1} report the corresponding Macro-F1 results. BIDETA remains the strongest overall method on both datasets; compared with accuracy, Macro-F1 reveals a more balanced advantage across classes, indicating greater robustness to class-dependent sensor shifts.

\begin{table}[!htbp]
\centering
\caption{\textbf{Per-sensor Macro-F1 on SITR with TVL.} Results (\%) are reported as mean $\pm$ sample standard deviation over three random seeds. Best results are in bold and second-best results are underlined.}
\label{tab:detail1f1}
\scriptsize
\setlength{\tabcolsep}{3pt}
\begin{adjustbox}{max width=\linewidth}
\begin{tabular}{lcccccccc}
\toprule
Method & Labels & Mini 1 & Mini 2 & Mini 3 & Mini 4 & Hex & Wedge & Mean \\
\midrule
Frozen backbone & 0\% & $2.16 \pm 0.23$ & $5.49 \pm 0.11$ & $2.27 \pm 0.11$ & $0.81 \pm 0.15$ & $0.73 \pm 0.00$ & $2.51 \pm 0.13$ & $2.33 \pm 0.01$ \\
Tip-Adapter & 1\% & $6.29 \pm 7.33$ & $22.92 \pm 6.58$ & $32.88 \pm 0.44$ & $5.58 \pm 0.49$ & $0.73 \pm 0.00$ & $2.55 \pm 0.09$ & $11.82 \pm 1.98$ \\
Tip-Adapter & 5\% & $18.74 \pm 0.16$ & $31.02 \pm 1.61$ & $41.26 \pm 0.83$ & $19.03 \pm 0.80$ & $0.74 \pm 0.00$ & $5.99 \pm 5.79$ & $19.46 \pm 0.66$ \\
Tip-Adapter & 10\% & $21.67 \pm 2.95$ & $34.30 \pm 0.65$ & $40.36 \pm 1.82$ & $23.51 \pm 1.63$ & $2.74 \pm 1.73$ & $26.07 \pm 3.28$ & $24.77 \pm 1.57$ \\
SimpleShot & 1\% & $31.31 \pm 3.60$ & $31.71 \pm 3.19$ & $48.72 \pm 0.63$ & $32.77 \pm 4.18$ & $\underline{32.65 \pm 2.74}$ & $\underline{40.07 \pm 0.91}$ & $\underline{36.20 \pm 0.88}$ \\
SimpleShot & 5\% & $39.67 \pm 2.49$ & $40.28 \pm 1.24$ & $57.92 \pm 1.69$ & $40.67 \pm 1.04$ & $\underline{36.68 \pm 0.95}$ & $45.18 \pm 1.06$ & $43.40 \pm 1.12$ \\
SimpleShot & 10\% & $37.79 \pm 1.47$ & $42.05 \pm 2.39$ & $57.29 \pm 2.28$ & $40.16 \pm 1.21$ & $\underline{36.21 \pm 1.30}$ & $\underline{46.70 \pm 0.48}$ & $43.37 \pm 0.74$ \\
LaplacianShot & 1\% & $27.49 \pm 2.45$ & $33.96 \pm 4.01$ & $53.22 \pm 3.00$ & $30.26 \pm 4.81$ & $32.53 \pm 3.67$ & $37.03 \pm 0.61$ & $35.75 \pm 0.96$ \\
LaplacianShot & 5\% & $39.79 \pm 3.25$ & $43.25 \pm 1.20$ & $60.15 \pm 2.82$ & $40.23 \pm 1.60$ & $36.03 \pm 1.44$ & $\underline{45.75 \pm 1.34}$ & $44.20 \pm 1.82$ \\
LaplacianShot & 10\% & $34.39 \pm 2.44$ & $41.85 \pm 3.43$ & $60.05 \pm 3.79$ & $37.29 \pm 0.75$ & $35.23 \pm 3.10$ & $44.07 \pm 1.78$ & $42.15 \pm 0.78$ \\
SITR-Calib & 1\% & $38.88 \pm 2.16$ & $\underline{36.35 \pm 1.68}$ & $54.40 \pm 1.46$ & $\underline{38.50 \pm 3.87}$ & $13.71 \pm 0.48$ & $35.11 \pm 3.87$ & $36.16 \pm 1.54$ \\
SITR-Calib & 5\% & $44.75 \pm 4.48$ & $55.76 \pm 0.77$ & $67.34 \pm 4.99$ & $40.44 \pm 1.90$ & $18.11 \pm 3.32$ & $39.18 \pm 2.87$ & $\underline{44.26 \pm 0.51}$ \\
SITR-Calib & 10\% & $43.58 \pm 0.24$ & $\underline{64.00 \pm 1.50}$ & $67.73 \pm 2.42$ & $40.41 \pm 2.59$ & $15.07 \pm 6.43$ & $40.99 \pm 2.23$ & $\underline{45.30 \pm 1.70}$ \\
AnyTouch Match & 1\% & $\underline{40.28 \pm 1.72}$ & $34.27 \pm 1.74$ & $54.27 \pm 1.58$ & $37.48 \pm 4.39$ & $3.61 \pm 0.43$ & $38.10 \pm 3.69$ & $34.67 \pm 1.41$ \\
AnyTouch Match & 5\% & $\underline{46.13 \pm 2.64}$ & $55.85 \pm 0.47$ & $\underline{69.61 \pm 2.46}$ & $\underline{41.65 \pm 1.24}$ & $3.96 \pm 0.13$ & $39.42 \pm 1.78$ & $42.77 \pm 0.30$ \\
AnyTouch Match & 10\% & $42.53 \pm 4.37$ & $63.11 \pm 1.09$ & $\underline{67.91 \pm 1.71}$ & $\underline{43.53 \pm 2.11}$ & $3.59 \pm 0.61$ & $41.68 \pm 4.23$ & $43.72 \pm 1.64$ \\
CTSRL CSM & 1\% & $38.24 \pm 1.26$ & $36.09 \pm 1.50$ & $\underline{54.68 \pm 2.48}$ & $35.56 \pm 4.72$ & $14.38 \pm 1.39$ & $34.58 \pm 1.43$ & $35.59 \pm 1.44$ \\
CTSRL CSM & 5\% & $43.99 \pm 1.67$ & $\underline{56.23 \pm 0.98}$ & $66.15 \pm 3.97$ & $40.41 \pm 2.23$ & $12.68 \pm 2.63$ & $34.33 \pm 2.94$ & $42.30 \pm 1.59$ \\
CTSRL CSM & 10\% & $\underline{44.33 \pm 1.11}$ & $62.64 \pm 1.58$ & $65.42 \pm 2.56$ & $40.75 \pm 2.33$ & $11.43 \pm 3.11$ & $35.53 \pm 4.13$ & $43.35 \pm 0.57$ \\
BIDETA & 1\% & $\mathbf{52.23 \pm 2.90}$ & $\mathbf{52.44 \pm 1.04}$ & $\mathbf{71.70 \pm 4.23}$ & $\mathbf{61.28 \pm 1.21}$ & $\mathbf{63.97 \pm 4.66}$ & $\mathbf{67.08 \pm 3.07}$ & $\mathbf{61.45 \pm 1.36}$ \\
BIDETA & 5\% & $\mathbf{77.12 \pm 1.58}$ & $\mathbf{78.27 \pm 1.03}$ & $\mathbf{87.80 \pm 2.48}$ & $\mathbf{80.82 \pm 1.19}$ & $\mathbf{81.93 \pm 1.98}$ & $\mathbf{91.69 \pm 1.73}$ & $\mathbf{82.94 \pm 0.79}$ \\
BIDETA & 10\% & $\mathbf{80.69 \pm 0.66}$ & $\mathbf{84.17 \pm 2.79}$ & $\mathbf{88.13 \pm 0.89}$ & $\mathbf{79.35 \pm 2.23}$ & $\mathbf{86.32 \pm 3.52}$ & $\mathbf{94.05 \pm 0.83}$ & $\mathbf{85.45 \pm 0.19}$ \\
\bottomrule
\end{tabular}
\end{adjustbox}
\end{table}

\begin{table}[!htbp]
\centering
\caption{\textbf{Per-sensor Macro-F1 on SITR with Sparsh.} Results (\%) are reported as mean $\pm$ sample standard deviation over three random seeds. Best results are in bold and second-best results are underlined.}
\label{tab:detail2f1}
\scriptsize
\setlength{\tabcolsep}{3pt}
\begin{adjustbox}{max width=\linewidth}
\begin{tabular}{lcccccccc}
\toprule
Method & Labels & Mini 1 & Mini 2 & Mini 3 & Mini 4 & Hex & Wedge & Mean \\
\midrule
Frozen backbone & 0\% & $0.74 \pm 0.00$ & $0.74 \pm 0.00$ & $0.74 \pm 0.00$ & $0.74 \pm 0.00$ & $0.76 \pm 0.04$ & $3.36 \pm 0.64$ & $1.18 \pm 0.11$ \\
Tip-Adapter & 1\% & $1.08 \pm 0.60$ & $25.36 \pm 3.13$ & $27.30 \pm 1.25$ & $0.74 \pm 0.00$ & $0.76 \pm 0.04$ & $3.46 \pm 0.60$ & $9.78 \pm 0.49$ \\
Tip-Adapter & 5\% & $16.58 \pm 0.37$ & $27.62 \pm 3.09$ & $32.34 \pm 1.62$ & $16.36 \pm 2.71$ & $0.78 \pm 0.07$ & $4.34 \pm 0.16$ & $16.34 \pm 1.12$ \\
Tip-Adapter & 10\% & $15.23 \pm 0.51$ & $30.77 \pm 2.41$ & $31.26 \pm 1.16$ & $14.33 \pm 1.00$ & $1.08 \pm 0.60$ & $4.68 \pm 0.20$ & $16.23 \pm 0.27$ \\
SimpleShot & 1\% & $23.80 \pm 2.41$ & $31.30 \pm 4.30$ & $35.04 \pm 3.40$ & $27.74 \pm 5.13$ & $\underline{28.68 \pm 2.80}$ & $\underline{36.97 \pm 2.28}$ & $30.59 \pm 2.71$ \\
SimpleShot & 5\% & $31.79 \pm 1.87$ & $34.96 \pm 2.79$ & $45.11 \pm 0.98$ & $35.02 \pm 3.05$ & $\underline{33.44 \pm 4.00}$ & $44.82 \pm 0.42$ & $37.52 \pm 0.80$ \\
SimpleShot & 10\% & $29.31 \pm 0.91$ & $39.91 \pm 3.24$ & $42.23 \pm 0.15$ & $34.46 \pm 1.81$ & $\underline{36.37 \pm 0.95}$ & $\underline{45.40 \pm 1.12}$ & $37.94 \pm 0.88$ \\
LaplacianShot & 1\% & $22.96 \pm 2.98$ & $\underline{34.88 \pm 6.70}$ & $36.23 \pm 6.12$ & $27.23 \pm 5.25$ & $27.96 \pm 5.89$ & $35.00 \pm 3.63$ & $\underline{30.71 \pm 3.82}$ \\
LaplacianShot & 5\% & $31.19 \pm 2.78$ & $\underline{38.59 \pm 3.01}$ & $46.49 \pm 1.24$ & $\underline{36.98 \pm 4.63}$ & $32.96 \pm 4.68$ & $\underline{45.76 \pm 1.14}$ & $\underline{38.66 \pm 1.20}$ \\
LaplacianShot & 10\% & $29.27 \pm 1.51$ & $\underline{42.89 \pm 4.45}$ & $42.21 \pm 1.64$ & $35.55 \pm 3.77$ & $34.55 \pm 0.81$ & $44.47 \pm 0.36$ & $\underline{38.16 \pm 0.91}$ \\
SITR-Calib & 1\% & $31.03 \pm 5.98$ & $32.01 \pm 2.13$ & $42.80 \pm 5.61$ & $26.67 \pm 1.31$ & $9.02 \pm 1.67$ & $24.69 \pm 0.84$ & $27.70 \pm 1.75$ \\
SITR-Calib & 5\% & $33.61 \pm 0.20$ & $33.78 \pm 3.45$ & $49.96 \pm 7.24$ & $32.92 \pm 1.57$ & $10.74 \pm 2.58$ & $26.82 \pm 0.41$ & $31.31 \pm 1.04$ \\
SITR-Calib & 10\% & $33.36 \pm 1.78$ & $35.72 \pm 2.07$ & $48.82 \pm 3.09$ & $34.31 \pm 1.19$ & $8.47 \pm 1.10$ & $24.44 \pm 2.72$ & $30.85 \pm 0.83$ \\
AnyTouch Match & 1\% & $\underline{35.09 \pm 5.25}$ & $34.05 \pm 2.28$ & $\underline{43.52 \pm 3.75}$ & $\underline{31.09 \pm 2.19}$ & $8.95 \pm 1.74$ & $26.76 \pm 2.81$ & $29.91 \pm 0.60$ \\
AnyTouch Match & 5\% & $\underline{37.40 \pm 0.90}$ & $38.43 \pm 2.37$ & $50.74 \pm 1.41$ & $35.10 \pm 2.67$ & $9.01 \pm 1.57$ & $24.59 \pm 4.96$ & $32.54 \pm 0.55$ \\
AnyTouch Match & 10\% & $\underline{35.87 \pm 2.18}$ & $40.44 \pm 1.29$ & $\underline{52.89 \pm 5.38}$ & $\underline{36.32 \pm 0.59}$ & $9.64 \pm 1.69$ & $24.68 \pm 4.97$ & $33.31 \pm 0.78$ \\
CTSRL CSM & 1\% & $32.84 \pm 2.40$ & $32.38 \pm 3.07$ & $43.11 \pm 3.45$ & $28.67 \pm 0.45$ & $14.27 \pm 1.30$ & $28.08 \pm 1.32$ & $29.89 \pm 1.12$ \\
CTSRL CSM & 5\% & $35.20 \pm 0.70$ & $36.42 \pm 1.47$ & $\underline{52.98 \pm 2.31}$ & $31.91 \pm 2.09$ & $11.98 \pm 2.64$ & $26.12 \pm 1.71$ & $32.43 \pm 0.35$ \\
CTSRL CSM & 10\% & $33.62 \pm 0.80$ & $36.01 \pm 2.03$ & $49.72 \pm 2.39$ & $32.99 \pm 1.09$ & $11.20 \pm 1.54$ & $27.61 \pm 1.26$ & $31.86 \pm 0.50$ \\
BIDETA & 1\% & $\mathbf{50.16 \pm 2.49}$ & $\mathbf{61.49 \pm 4.48}$ & $\mathbf{62.41 \pm 5.01}$ & $\mathbf{55.37 \pm 3.25}$ & $\mathbf{65.14 \pm 4.44}$ & $\mathbf{72.42 \pm 2.82}$ & $\mathbf{61.16 \pm 1.79}$ \\
BIDETA & 5\% & $\mathbf{66.04 \pm 2.45}$ & $\mathbf{93.45 \pm 1.17}$ & $\mathbf{79.73 \pm 1.90}$ & $\mathbf{81.70 \pm 1.31}$ & $\mathbf{89.05 \pm 1.16}$ & $\mathbf{90.47 \pm 1.70}$ & $\mathbf{83.41 \pm 0.79}$ \\
BIDETA & 10\% & $\mathbf{71.21 \pm 2.59}$ & $\mathbf{95.80 \pm 0.09}$ & $\mathbf{84.25 \pm 0.82}$ & $\mathbf{85.91 \pm 1.26}$ & $\mathbf{92.84 \pm 0.80}$ & $\mathbf{91.20 \pm 1.18}$ & $\mathbf{86.87 \pm 0.58}$ \\
\bottomrule
\end{tabular}
\end{adjustbox}
\end{table}

\begin{table}[!htbp]
\centering
\caption{\textbf{Per-sensor Macro-F1 on TacVerse Shape with TVL.} Results (\%) are reported as mean $\pm$ sample standard deviation over three random seeds. Best results are in bold and second-best results are underlined. SITR-Calib is unavailable because standard calibration images are absent.}
\label{tab:detail3f1}
\scriptsize
\setlength{\tabcolsep}{3pt}
\begin{adjustbox}{max width=\linewidth}
\begin{tabular}{lccccccc}
\toprule
Method & Labels & MagicGripper & MagicTac & TacTip & ViTac & ViTacTip & Mean \\
\midrule
Frozen backbone & 0\% & $6.91 \pm 0.60$ & $2.51 \pm 0.35$ & $3.89 \pm 0.72$ & $2.22 \pm 0.00$ & $2.22 \pm 0.00$ & $3.55 \pm 0.13$ \\
Tip-Adapter & 1\% & $21.20 \pm 6.34$ & $23.88 \pm 15.70$ & $3.95 \pm 0.65$ & $15.72 \pm 12.21$ & $12.29 \pm 3.31$ & $15.41 \pm 4.29$ \\
Tip-Adapter & 5\% & $38.41 \pm 3.42$ & $46.44 \pm 0.51$ & $9.47 \pm 8.10$ & $33.06 \pm 1.40$ & $34.99 \pm 0.38$ & $32.47 \pm 1.71$ \\
Tip-Adapter & 10\% & $39.73 \pm 4.36$ & $46.48 \pm 0.44$ & $20.97 \pm 0.52$ & $34.40 \pm 3.49$ & $59.45 \pm 1.23$ & $40.21 \pm 0.48$ \\
SimpleShot & 1\% & $30.86 \pm 6.58$ & $43.25 \pm 5.01$ & $\underline{26.11 \pm 3.10}$ & $\underline{51.00 \pm 7.27}$ & $\underline{57.09 \pm 7.62}$ & $\underline{41.66 \pm 2.31}$ \\
SimpleShot & 5\% & $40.03 \pm 3.05$ & $44.70 \pm 0.64$ & $\underline{29.40 \pm 1.59}$ & $\underline{61.37 \pm 3.13}$ & $\underline{68.62 \pm 3.98}$ & $48.82 \pm 1.67$ \\
SimpleShot & 10\% & $46.83 \pm 3.08$ & $45.62 \pm 1.86$ & $\underline{31.90 \pm 0.67}$ & $62.01 \pm 3.42$ & $68.34 \pm 3.09$ & $50.94 \pm 0.81$ \\
LaplacianShot & 1\% & $\underline{31.59 \pm 8.16}$ & $43.70 \pm 3.28$ & $19.57 \pm 6.81$ & $43.57 \pm 6.93$ & $49.56 \pm 3.49$ & $37.60 \pm 4.02$ \\
LaplacianShot & 5\% & $\underline{41.05 \pm 5.03}$ & $43.13 \pm 2.32$ & $23.74 \pm 4.55$ & $59.28 \pm 5.49$ & $58.25 \pm 5.91$ & $45.09 \pm 1.82$ \\
LaplacianShot & 10\% & $48.12 \pm 3.50$ & $44.67 \pm 1.86$ & $25.62 \pm 1.25$ & $57.82 \pm 4.58$ & $55.61 \pm 6.29$ & $46.37 \pm 1.78$ \\
SITR-Calib & 1\% & -- & -- & -- & -- & -- & -- \\
SITR-Calib & 5\% & -- & -- & -- & -- & -- & -- \\
SITR-Calib & 10\% & -- & -- & -- & -- & -- & -- \\
AnyTouch Match & 1\% & $27.54 \pm 3.23$ & $50.96 \pm 12.71$ & $17.02 \pm 1.82$ & $48.27 \pm 8.14$ & $52.97 \pm 8.30$ & $39.35 \pm 1.66$ \\
AnyTouch Match & 5\% & $39.04 \pm 9.14$ & $67.45 \pm 3.64$ & $19.46 \pm 1.39$ & $55.47 \pm 9.09$ & $61.81 \pm 5.03$ & $48.65 \pm 1.36$ \\
AnyTouch Match & 10\% & $\underline{49.34 \pm 2.41}$ & $68.30 \pm 3.57$ & $18.62 \pm 1.18$ & $\underline{65.39 \pm 0.66}$ & $65.60 \pm 3.38$ & $53.45 \pm 0.35$ \\
CTSRL CSM & 1\% & $28.19 \pm 2.03$ & $\underline{52.24 \pm 16.38}$ & $18.79 \pm 3.60$ & $49.85 \pm 7.48$ & $52.37 \pm 8.66$ & $40.29 \pm 1.28$ \\
CTSRL CSM & 5\% & $40.23 \pm 8.47$ & $\underline{70.22 \pm 4.02}$ & $22.41 \pm 3.50$ & $56.99 \pm 1.85$ & $64.59 \pm 6.66$ & $\underline{50.89 \pm 4.24}$ \\
CTSRL CSM & 10\% & $49.03 \pm 3.48$ & $\underline{69.68 \pm 1.48}$ & $21.20 \pm 1.44$ & $61.80 \pm 1.65$ & $\underline{70.24 \pm 1.34}$ & $\underline{54.39 \pm 0.66}$ \\
BIDETA & 1\% & $\mathbf{36.53 \pm 3.79}$ & $\mathbf{71.47 \pm 8.13}$ & $\mathbf{35.67 \pm 2.94}$ & $\mathbf{71.06 \pm 2.09}$ & $\mathbf{67.32 \pm 4.00}$ & $\mathbf{56.41 \pm 1.84}$ \\
BIDETA & 5\% & $\mathbf{51.88 \pm 5.09}$ & $\mathbf{99.44 \pm 0.87}$ & $\mathbf{41.12 \pm 1.28}$ & $\mathbf{95.77 \pm 0.68}$ & $\mathbf{86.93 \pm 4.44}$ & $\mathbf{75.03 \pm 1.98}$ \\
BIDETA & 10\% & $\mathbf{56.32 \pm 6.90}$ & $\mathbf{99.48 \pm 0.71}$ & $\mathbf{50.39 \pm 3.48}$ & $\mathbf{97.49 \pm 1.09}$ & $\mathbf{90.22 \pm 1.30}$ & $\mathbf{78.78 \pm 1.15}$ \\
\bottomrule
\end{tabular}
\end{adjustbox}
\end{table}

\begin{table}[!htbp]
\centering
\caption{\textbf{Per-sensor Macro-F1 on TacVerse Shape with Sparsh.} Results (\%) are reported as mean $\pm$ sample standard deviation over three random seeds. Best results are in bold and second-best results are underlined. SITR-Calib is unavailable because standard calibration images are absent.}
\label{tab:detail4f1}
\scriptsize
\setlength{\tabcolsep}{3pt}
\begin{adjustbox}{max width=\linewidth}
\begin{tabular}{lccccccc}
\toprule
Method & Labels & MagicGripper & MagicTac & TacTip & ViTac & ViTacTip & Mean \\
\midrule
Frozen backbone & 0\% & $5.47 \pm 0.45$ & $2.22 \pm 0.00$ & $2.22 \pm 0.00$ & $6.58 \pm 0.80$ & $6.68 \pm 1.00$ & $4.64 \pm 0.19$ \\
Tip-Adapter & 1\% & $7.03 \pm 1.78$ & $2.22 \pm 0.00$ & $2.22 \pm 0.00$ & $6.68 \pm 0.74$ & $6.70 \pm 1.02$ & $4.97 \pm 0.52$ \\
Tip-Adapter & 5\% & $36.49 \pm 2.63$ & $2.22 \pm 0.00$ & $2.22 \pm 0.00$ & $16.32 \pm 0.43$ & $16.99 \pm 0.38$ & $14.85 \pm 0.66$ \\
Tip-Adapter & 10\% & $37.81 \pm 1.37$ & $56.27 \pm 3.17$ & $2.22 \pm 0.00$ & $16.92 \pm 0.24$ & $18.34 \pm 0.54$ & $26.31 \pm 0.54$ \\
SimpleShot & 1\% & $34.68 \pm 1.33$ & $\underline{45.42 \pm 3.84}$ & $\underline{35.36 \pm 4.27}$ & $54.47 \pm 6.34$ & $44.60 \pm 4.05$ & $\underline{42.91 \pm 0.49}$ \\
SimpleShot & 5\% & $44.55 \pm 2.74$ & $54.89 \pm 2.81$ & $\underline{37.79 \pm 0.90}$ & $\underline{66.01 \pm 2.53}$ & $48.71 \pm 1.80$ & $\underline{50.39 \pm 0.26}$ \\
SimpleShot & 10\% & $43.95 \pm 3.53$ & $\underline{58.55 \pm 2.83}$ & $\underline{43.49 \pm 2.88}$ & $\underline{68.17 \pm 2.01}$ & $\underline{52.38 \pm 1.48}$ & $\underline{53.31 \pm 1.25}$ \\
LaplacianShot & 1\% & $32.50 \pm 2.81$ & $43.16 \pm 4.91$ & $30.67 \pm 3.43$ & $49.51 \pm 6.99$ & $37.99 \pm 9.04$ & $38.77 \pm 2.84$ \\
LaplacianShot & 5\% & $43.62 \pm 3.44$ & $\underline{55.16 \pm 2.42}$ & $36.15 \pm 5.13$ & $60.20 \pm 0.33$ & $37.75 \pm 8.50$ & $46.57 \pm 1.62$ \\
LaplacianShot & 10\% & $41.26 \pm 2.54$ & $56.24 \pm 3.20$ & $36.15 \pm 3.23$ & $63.94 \pm 5.98$ & $34.55 \pm 4.16$ & $46.43 \pm 2.14$ \\
SITR-Calib & 1\% & -- & -- & -- & -- & -- & -- \\
SITR-Calib & 5\% & -- & -- & -- & -- & -- & -- \\
SITR-Calib & 10\% & -- & -- & -- & -- & -- & -- \\
AnyTouch Match & 1\% & $37.72 \pm 2.87$ & $43.10 \pm 7.27$ & $2.22 \pm 0.00$ & $54.40 \pm 4.61$ & $46.12 \pm 1.92$ & $36.71 \pm 2.29$ \\
AnyTouch Match & 5\% & $\underline{57.65 \pm 2.90}$ & $50.32 \pm 1.26$ & $2.22 \pm 0.00$ & $58.69 \pm 1.34$ & $\underline{55.58 \pm 2.65}$ & $44.89 \pm 0.83$ \\
AnyTouch Match & 10\% & $\underline{57.94 \pm 2.41}$ & $55.03 \pm 4.87$ & $2.22 \pm 0.00$ & $63.61 \pm 4.27$ & $49.98 \pm 4.48$ & $45.76 \pm 2.00$ \\
CTSRL CSM & 1\% & $\underline{37.83 \pm 6.49}$ & $42.01 \pm 6.24$ & $5.54 \pm 1.11$ & $\mathbf{58.70 \pm 4.67}$ & $\underline{50.44 \pm 2.78}$ & $38.91 \pm 2.21$ \\
CTSRL CSM & 5\% & $52.67 \pm 5.77$ & $48.53 \pm 2.66$ & $5.70 \pm 3.34$ & $65.23 \pm 0.62$ & $52.86 \pm 1.00$ & $45.00 \pm 1.79$ \\
CTSRL CSM & 10\% & $54.50 \pm 3.45$ & $49.62 \pm 4.58$ & $6.03 \pm 1.10$ & $65.63 \pm 1.96$ & $51.98 \pm 2.43$ & $45.55 \pm 2.04$ \\
BIDETA & 1\% & $\mathbf{48.45 \pm 8.25}$ & $\mathbf{88.95 \pm 5.31}$ & $\mathbf{37.75 \pm 2.06}$ & $\underline{56.30 \pm 3.04}$ & $\mathbf{59.17 \pm 6.55}$ & $\mathbf{58.12 \pm 2.49}$ \\
BIDETA & 5\% & $\mathbf{81.81 \pm 7.34}$ & $\mathbf{100.00 \pm 0.00}$ & $\mathbf{50.84 \pm 1.50}$ & $\mathbf{86.77 \pm 3.44}$ & $\mathbf{80.14 \pm 3.75}$ & $\mathbf{79.91 \pm 0.16}$ \\
BIDETA & 10\% & $\mathbf{89.67 \pm 3.41}$ & $\mathbf{100.00 \pm 0.00}$ & $\mathbf{61.92 \pm 2.19}$ & $\mathbf{86.88 \pm 1.88}$ & $\mathbf{77.14 \pm 5.91}$ & $\mathbf{83.12 \pm 1.19}$ \\
\bottomrule
\end{tabular}
\end{adjustbox}
\end{table}

\subsection{Unified Shot Evaluation}
\label{app:unified_shots}
Each shot denotes the number of labeled support images per target class under the unified 2-, 5-, and 10-shot protocols. Within each random seed, the support sets are nested as $2\text{-shot}\subset5\text{-shot}\subset10\text{-shot}$. Tables~\ref{tab:sitr_tvl_shots}--\ref{tab:tacverse_sparsh_shots} report the corresponding classification results.
\begin{table}[!htbp]
\centering
\caption{\textbf{Unified-shot classification on SITR with the frozen TVL ViT-Small backbone.} Accuracy and Macro-F1 (\%) are averaged over the five target sensors excluded from hyperparameter development and reported as mean $\pm$ sample standard deviation over three random seeds. Best results are in bold and second-best results are underlined.}
\label{tab:sitr_tvl_shots}
\scriptsize
\setlength{\tabcolsep}{2.8pt}
\renewcommand{\arraystretch}{1.08}
\begin{adjustbox}{max width=\linewidth}
\begin{tabular}{lcccccc}
\toprule 
& \multicolumn{2}{c}{2-shot} & \multicolumn{2}{c}{5-shot} & \multicolumn{2}{c}{10-shot} \\
\cmidrule(lr){2-3}\cmidrule(lr){4-5}\cmidrule(lr){6-7}
Method & Acc. & Macro-F1 & Acc. & Macro-F1 & Acc. & Macro-F1 \\
\midrule
Frozen + source head & $7.77{\pm}0.18$ & $2.34{\pm}0.03$ & $7.77{\pm}0.18$ & $2.34{\pm}0.03$ & $7.77{\pm}0.18$ & $2.34{\pm}0.03$ \\
Tip-Adapter & $8.20{\pm}0.70$ & $2.80{\pm}0.79$ & $11.53{\pm}1.18$ & $5.52{\pm}1.19$ & $14.97{\pm}2.07$ & $9.47{\pm}2.12$ \\
SimpleShot & $28.49{\pm}2.13$ & $\underline{26.36{\pm}1.86}$ & $33.49{\pm}1.35$ & $\underline{31.58{\pm}1.20}$ & $36.68{\pm}1.26$ & $35.05{\pm}1.04$ \\
LaplacianShot & $\underline{28.55{\pm}3.57}$ & $25.38{\pm}3.23$ & $\underline{34.33{\pm}0.74}$ & $31.02{\pm}0.58$ & $\underline{37.77{\pm}1.26}$ & $34.70{\pm}1.18$ \\
SITR-Calib & $26.58{\pm}1.77$ & $24.95{\pm}2.05$ & $32.45{\pm}2.47$ & $31.03{\pm}2.84$ & $37.15{\pm}2.40$ & $\underline{35.06{\pm}2.61}$ \\
AnyTouch Match & $24.92{\pm}1.53$ & $23.22{\pm}1.89$ & $29.35{\pm}1.64$ & $28.38{\pm}1.69$ & $33.30{\pm}0.58$ & $31.61{\pm}0.37$ \\
CTSRL CSM & $25.95{\pm}1.27$ & $24.22{\pm}1.37$ & $30.57{\pm}2.13$ & $29.30{\pm}2.01$ & $34.33{\pm}1.22$ & $32.56{\pm}1.41$ \\
BIDETA & $\mathbf{36.40{\pm}1.46}$ & $\mathbf{35.00{\pm}1.38}$ & $\mathbf{54.90{\pm}2.91}$ & $\mathbf{53.96{\pm}3.10}$ & $\mathbf{65.83{\pm}1.84}$ & $\mathbf{64.85{\pm}1.67}$ \\
\bottomrule
\end{tabular}
\end{adjustbox}
\end{table}
 
\begin{table}[!htbp]
\centering
\caption{\textbf{Unified-shot classification on SITR with the frozen Sparsh-DINO Small backbone.} Accuracy and Macro-F1 (\%) are averaged over the five target sensors excluded from hyperparameter development and reported as mean $\pm$ sample standard deviation over three random seeds. Best results are in bold and second-best results are underlined.}
\label{tab:sitr_sparsh_shots}
\scriptsize
\setlength{\tabcolsep}{2.8pt}
\renewcommand{\arraystretch}{1.08}
\begin{adjustbox}{max width=\linewidth}
\begin{tabular}{lcccccc}
\toprule
& \multicolumn{2}{c}{2-shot} & \multicolumn{2}{c}{5-shot} & \multicolumn{2}{c}{10-shot} \\
\cmidrule(lr){2-3}\cmidrule(lr){4-5}\cmidrule(lr){6-7}
Method & Acc. & Macro-F1 & Acc. & Macro-F1 & Acc. & Macro-F1 \\
\midrule
Frozen + source head & $6.98{\pm}0.40$ & $1.26{\pm}0.14$ & $6.98{\pm}0.40$ & $1.26{\pm}0.14$ & $6.98{\pm}0.40$ & $1.26{\pm}0.14$ \\
Tip-Adapter & $6.99{\pm}0.40$ & $1.27{\pm}0.14$ & $9.69{\pm}1.20$ & $4.05{\pm}1.56$ & $12.57{\pm}0.42$ & $7.11{\pm}0.33$ \\
SimpleShot & $24.39{\pm}0.62$ & $23.11{\pm}0.96$ & $30.87{\pm}1.54$ & $29.83{\pm}1.78$ & $34.45{\pm}0.98$ & $33.41{\pm}0.98$ \\
LaplacianShot & $\underline{25.17{\pm}0.71}$ & $\underline{23.39{\pm}1.13}$ & $\underline{32.26{\pm}1.46}$ & $\underline{30.58{\pm}1.64}$ & $\underline{34.94{\pm}1.51}$ & $\underline{33.44{\pm}1.47}$ \\
SITR-Calib & $21.79{\pm}1.66$ & $19.41{\pm}1.38$ & $29.20{\pm}0.41$ & $26.21{\pm}0.89$ & $29.08{\pm}1.85$ & $26.42{\pm}1.57$ \\
AnyTouch Match & $22.66{\pm}1.22$ & $20.58{\pm}0.72$ & $29.23{\pm}0.33$ & $27.14{\pm}0.35$ & $29.81{\pm}1.34$ & $27.31{\pm}1.78$ \\
CTSRL CSM & $23.17{\pm}0.67$ & $21.34{\pm}1.29$ & $29.67{\pm}0.35$ & $27.29{\pm}0.67$ & $30.55{\pm}1.22$ & $28.43{\pm}1.22$ \\
BIDETA & $\mathbf{35.15{\pm}0.55}$ & $\mathbf{34.54{\pm}0.45}$ & $\mathbf{57.52{\pm}0.54}$ & $\mathbf{56.87{\pm}0.65}$ & $\mathbf{70.06{\pm}0.48}$ & $\mathbf{69.58{\pm}0.72}$ \\
\bottomrule
\end{tabular}
\end{adjustbox}
\end{table}

\begin{table}[!htbp]
\centering
\caption{\textbf{Unified-shot classification on TacVerse Shape with the frozen TVL ViT-Small backbone.} Accuracy and Macro-F1 (\%) are averaged over five target sensors and reported as mean $\pm$ sample standard deviation over three random seeds. SITR-Calib is unavailable because standard calibration images are absent. Best results are in bold and second-best results are underlined.}
\label{tab:tacverse_tvl_shots}
\scriptsize
\setlength{\tabcolsep}{2.8pt}
\renewcommand{\arraystretch}{1.08}
\begin{adjustbox}{max width=\linewidth}
\begin{tabular}{lcccccc}
\toprule
& \multicolumn{2}{c}{2-shot} & \multicolumn{2}{c}{5-shot} & \multicolumn{2}{c}{10-shot} \\
\cmidrule(lr){2-3}\cmidrule(lr){4-5}\cmidrule(lr){6-7}
Method & Acc. & Macro-F1 & Acc. & Macro-F1 & Acc. & Macro-F1 \\
\midrule
Frozen + source head & $9.56{\pm}0.48$ & $3.55{\pm}0.13$ & $9.56{\pm}0.48$ & $3.55{\pm}0.13$ & $9.56{\pm}0.48$ & $3.55{\pm}0.13$ \\
Tip-Adapter & $15.47{\pm}3.11$ & $9.20{\pm}3.18$ & $28.64{\pm}1.87$ & $22.20{\pm}1.89$ & $36.35{\pm}1.15$ & $29.93{\pm}1.36$ \\
SimpleShot & $40.61{\pm}1.81$ & $\underline{38.21{\pm}2.13}$ & $\underline{46.07{\pm}1.76}$ & $\underline{43.71{\pm}1.77}$ & $49.41{\pm}2.44$ & $\underline{47.16{\pm}2.68}$ \\
LaplacianShot & $\underline{40.96{\pm}2.88}$ & $34.78{\pm}3.99$ & $43.90{\pm}2.91$ & $37.65{\pm}3.58$ & $46.82{\pm}2.22$ & $41.11{\pm}2.93$ \\
AnyTouch Match & $36.66{\pm}1.89$ & $35.07{\pm}1.47$ & $43.76{\pm}2.32$ & $41.19{\pm}2.38$ & $\underline{50.07{\pm}1.64}$ & $47.09{\pm}1.58$ \\
CTSRL CSM & $36.79{\pm}2.17$ & $35.12{\pm}1.42$ & $45.41{\pm}2.25$ & $43.15{\pm}2.54$ & $49.77{\pm}2.20$ & $46.40{\pm}2.62$ \\
BIDETA & $\mathbf{47.90{\pm}1.08}$ & $\mathbf{46.41{\pm}1.24}$ & $\mathbf{67.04{\pm}2.80}$ & $\mathbf{66.42{\pm}2.77}$ & $\mathbf{72.99{\pm}0.92}$ & $\mathbf{72.23{\pm}1.15}$ \\
\bottomrule
\end{tabular}
\end{adjustbox}
\end{table}

\begin{table}[!htbp]
\centering
\caption{\textbf{Unified-shot classification on TacVerse Shape with the frozen Sparsh-DINO Small backbone.} Accuracy and Macro-F1 (\%) are averaged over five target sensors and reported as mean $\pm$ sample standard deviation over three random seeds. SITR-Calib is unavailable because standard calibration images are absent. Best results are in bold and second-best results are underlined.}
\label{tab:tacverse_sparsh_shots}
\scriptsize
\setlength{\tabcolsep}{2.8pt}
\renewcommand{\arraystretch}{1.08}
\begin{adjustbox}{max width=\linewidth}
\begin{tabular}{lcccccc}
\toprule
& \multicolumn{2}{c}{2-shot} & \multicolumn{2}{c}{5-shot} & \multicolumn{2}{c}{10-shot} \\
\cmidrule(lr){2-3}\cmidrule(lr){4-5}\cmidrule(lr){6-7}
Method & Acc. & Macro-F1 & Acc. & Macro-F1 & Acc. & Macro-F1 \\
\midrule
Frozen + source head & $13.67{\pm}0.47$ & $4.64{\pm}0.19$ & $13.67{\pm}0.47$ & $4.64{\pm}0.19$ & $13.67{\pm}0.47$ & $4.64{\pm}0.19$ \\
Tip-Adapter & $14.28{\pm}1.17$ & $5.00{\pm}0.69$ & $21.21{\pm}0.43$ & $10.26{\pm}0.84$ & $23.66{\pm}2.07$ & $12.91{\pm}2.49$ \\
SimpleShot & $\underline{41.33{\pm}1.43}$ & $\underline{38.70{\pm}3.28}$ & $\underline{51.21{\pm}2.44}$ & $\underline{48.23{\pm}4.01}$ & $\underline{53.20{\pm}0.89}$ & $\underline{50.15{\pm}1.41}$ \\
LaplacianShot & $39.83{\pm}3.21$ & $34.89{\pm}5.21$ & $50.50{\pm}4.01$ & $45.46{\pm}5.91$ & $50.85{\pm}2.21$ & $44.86{\pm}3.66$ \\
AnyTouch Match & $37.01{\pm}3.71$ & $32.85{\pm}4.12$ & $43.79{\pm}1.69$ & $39.04{\pm}2.53$ & $48.01{\pm}0.37$ & $43.82{\pm}0.80$ \\
CTSRL CSM & $39.30{\pm}3.42$ & $35.06{\pm}4.05$ & $45.44{\pm}1.40$ & $41.27{\pm}2.35$ & $49.27{\pm}0.84$ & $44.49{\pm}1.51$ \\
BIDETA & $\mathbf{48.55{\pm}4.73}$ & $\mathbf{47.52{\pm}5.72}$ & $\mathbf{70.73{\pm}1.62}$ & $\mathbf{69.46{\pm}1.56}$ & $\mathbf{77.59{\pm}1.67}$ & $\mathbf{76.78{\pm}1.86}$ \\
\bottomrule
\end{tabular}
\end{adjustbox}
\end{table}

\subsection{Class-wise prediction structure}
\begin{figure}[!htbp]
\centering
\includegraphics[width=.94\linewidth]{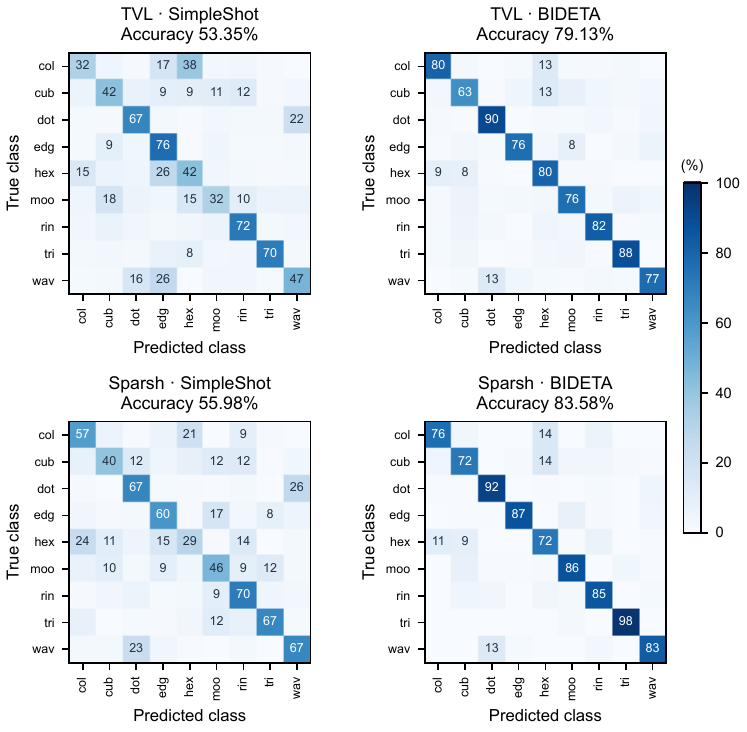}
\caption{\textbf{TacVerse class predictions with 10\% support.} Row-normalized confusion matrices compare SimpleShot and BIDETA on identical queries. Each matrix aggregates 13,500 predictions from five target sensors, three random seeds, and 900 queries per target.}
\label{fig:tacverseconfusion}
\end{figure}

Figure~\ref{fig:tacverseconfusion} compares BIDETA with SimpleShot on the same TacVerse queries using 10\% support. BIDETA provides stronger class discrimination and cross-class balance, with predictions concentrated more clearly along the diagonal than SimpleShot, producing more accurate and stable recognition.

\begin{table}[!htbp]
\centering
\caption{\textbf{TacQuad identity-ranking results with TVL.} MRR and R@1 (\%) are reported as mean $\pm$ sample standard deviation over three random seeds. Best results are in bold and second-best results are underlined.}
\label{tab:ranking5}
\small
\setlength{\tabcolsep}{3pt}
\begin{adjustbox}{max width=\linewidth}
\begin{tabular}{lcccccc}
\toprule
Method & R@1 10\% & MRR 10\% & R@1 20\% & MRR 20\% & R@1 30\% & MRR 30\% \\
\midrule
Frozen backbone & $2.46 \pm 0.00$ & $8.76 \pm 0.00$ & $2.46 \pm 0.00$ & $8.76 \pm 0.00$ & $2.46 \pm 0.00$ & $8.76 \pm 0.00$ \\
Tip-Adapter & $24.18 \pm 3.93$ & $36.68 \pm 3.70$ & $34.90 \pm 11.12$ & $47.80 \pm 12.36$ & $41.29 \pm 11.62$ & $53.80 \pm 9.55$ \\
SimpleShot & $\underline{48.07 \pm 1.57}$ & $\underline{59.05 \pm 0.78}$ & $53.42 \pm 2.91$ & $64.65 \pm 1.65$ & $54.54 \pm 1.95$ & $66.57 \pm 1.06$ \\
LaplacianShot & $24.93 \pm 5.86$ & $45.11 \pm 3.03$ & $27.08 \pm 2.95$ & $48.78 \pm 2.27$ & $23.96 \pm 1.90$ & $47.64 \pm 1.44$ \\
SITR-Support & $45.39 \pm 2.91$ & $57.94 \pm 2.00$ & $52.16 \pm 0.52$ & $64.54 \pm 0.38$ & $52.38 \pm 1.52$ & $65.49 \pm 1.42$ \\
AnyTouch Match & $45.76 \pm 0.97$ & $58.29 \pm 1.37$ & $\underline{54.09 \pm 4.15}$ & $\underline{66.15 \pm 2.49}$ & $\underline{56.32 \pm 1.49}$ & $\underline{68.49 \pm 1.10}$ \\
CTSRL CSM & $45.39 \pm 3.19$ & $57.49 \pm 2.16$ & $51.19 \pm 1.79$ & $63.88 \pm 0.88$ & $52.16 \pm 2.34$ & $65.38 \pm 1.35$ \\
BIDETA & $\mathbf{50.97 \pm 1.01}$ & $\mathbf{61.16 \pm 0.95}$ & $\mathbf{69.35 \pm 0.93}$ & $\mathbf{76.98 \pm 0.20}$ & $\mathbf{74.85 \pm 2.28}$ & $\mathbf{80.90 \pm 1.89}$ \\
\bottomrule
\end{tabular}
\end{adjustbox}
\end{table}

\begin{table}[!htbp]
\centering
\caption{\textbf{TacQuad identity-ranking results with Sparsh.} MRR and R@1 (\%) are reported as mean $\pm$ sample standard deviation over three random seeds. Best results are in bold and second-best results are underlined.}
\label{tab:ranking6}
\small
\setlength{\tabcolsep}{3pt}
\begin{adjustbox}{max width=\linewidth}
\begin{tabular}{lcccccc}
\toprule
Method & R@1 10\% & MRR 10\% & R@1 20\% & MRR 20\% & R@1 30\% & MRR 30\% \\
\midrule
Frozen backbone & $1.34 \pm 0.00$ & $10.01 \pm 0.00$ & $1.34 \pm 0.00$ & $10.01 \pm 0.00$ & $1.34 \pm 0.00$ & $10.01 \pm 0.00$ \\
Tip-Adapter & $36.90 \pm 0.34$ & $45.80 \pm 0.69$ & $51.04 \pm 1.23$ & $61.42 \pm 1.17$ & $54.39 \pm 1.23$ & $64.57 \pm 0.73$ \\
SimpleShot & $51.26 \pm 0.93$ & $60.36 \pm 0.85$ & $56.55 \pm 1.61$ & $65.83 \pm 1.20$ & $60.27 \pm 0.80$ & $68.75 \pm 0.78$ \\
LaplacianShot & $21.65 \pm 3.72$ & $43.32 \pm 3.10$ & $18.15 \pm 3.16$ & $43.61 \pm 2.61$ & $20.16 \pm 0.85$ & $45.82 \pm 0.88$ \\
SITR-Support & $51.79 \pm 1.83$ & $61.49 \pm 1.62$ & $\underline{57.66 \pm 0.78}$ & $67.90 \pm 0.63$ & $60.49 \pm 0.59$ & $70.72 \pm 0.44$ \\
AnyTouch Match & $\underline{51.86 \pm 0.68}$ & $\underline{61.73 \pm 0.93}$ & $57.07 \pm 1.92$ & $\underline{67.93 \pm 0.85}$ & $\underline{60.64 \pm 1.36}$ & $\underline{71.00 \pm 0.51}$ \\
CTSRL CSM & $48.29 \pm 1.01$ & $58.88 \pm 0.78$ & $53.79 \pm 1.36$ & $64.99 \pm 1.11$ & $57.37 \pm 0.67$ & $68.14 \pm 0.23$ \\
BIDETA & $\mathbf{62.05 \pm 1.36}$ & $\mathbf{69.28 \pm 0.72}$ & $\mathbf{74.85 \pm 2.83}$ & $\mathbf{80.82 \pm 2.67}$ & $\mathbf{80.73 \pm 2.19}$ & $\mathbf{85.86 \pm 1.80}$ \\
\bottomrule
\end{tabular}
\end{adjustbox}
\end{table}

\section{Additional identity-ranking results}
\label{app:ranking}
Tables~\ref{tab:ranking5} and~\ref{tab:ranking6} present identity-ranking results on TacQuad with different pretrained backbones. R@1 is the percentage of queries for which the correct identity ranks first and is equivalent to closed-set identity classification accuracy. Compared with the other methods, BIDETA consistently achieves higher first-rank accuracy and MRR across both backbones and all support budgets, demonstrating stronger cross-sensor identity discrimination.

\section{Complete component evidence}
\label{app:ablation}
\begin{table}[!htbp]
\centering
\caption{\textbf{Complete component ablation on SITR.} Macro-F1 (\%) is reported as mean $\pm$ sample standard deviation over three random seeds. Best results are in bold and second-best results are underlined.}
\label{tab:fullablationf1}
\small
\setlength{\tabcolsep}{3pt}
\begin{adjustbox}{max width=\linewidth}
\begin{tabular}{llcccccc}
\toprule
Type & Variant & TVL 1\% & TVL 5\% & TVL 10\% & Sparsh 1\% & Sparsh 5\% & Sparsh 10\% \\
\midrule
\multirow{3}{*}{Memory} & View 1 only & $49.61 \pm 0.97$ & $71.88 \pm 0.94$ & $75.25 \pm 0.28$ & $50.18 \pm 3.23$ & $74.41 \pm 0.14$ & $76.89 \pm 0.63$ \\
& View 2 only & $48.62 \pm 1.79$ & $70.26 \pm 0.92$ & $75.45 \pm 0.13$ & $53.06 \pm 1.63$ & $70.68 \pm 0.19$ & $76.91 \pm 0.86$ \\
& Dual-view memory & $50.84 \pm 0.98$ & $72.13 \pm 0.95$ & $75.48 \pm 0.21$ & $51.38 \pm 3.11$ & $73.84 \pm 0.13$ & $77.16 \pm 0.83$ \\
\midrule
\multirow{3}{*}{Geometry} & + Consensus recurrence & $55.69 \pm 1.06$ & $76.76 \pm 0.61$ & $79.72 \pm 0.22$ & $54.36 \pm 2.65$ & $78.18 \pm 0.72$ & $80.43 \pm 0.73$ \\
& + Support standardization & $56.35 \pm 1.09$ & $77.37 \pm 0.62$ & $80.14 \pm 0.23$ & $55.14 \pm 2.56$ & $78.78 \pm 0.72$ & $81.12 \pm 0.65$ \\
& + Spectral normalization & $58.13 \pm 1.26$ & $80.98 \pm 0.78$ & $82.40 \pm 0.08$ & $58.16 \pm 2.50$ & $81.33 \pm 0.95$ & $83.84 \pm 0.83$ \\
\midrule
\multirow{5}{*}{Fusion and gate} & Arithmetic graph fusion & $60.96 \pm 1.39$ & $82.55 \pm 0.76$ & $85.43 \pm 0.23$ & $60.88 \pm 1.82$ & $83.19 \pm 0.76$ & $86.65 \pm 0.64$ \\
& Without spectral normalization & $57.02 \pm 1.14$ & $75.43 \pm 0.52$ & $81.10 \pm 0.13$ & $55.27 \pm 2.20$ & $77.16 \pm 0.42$ & $81.98 \pm 0.73$ \\
& Disagreement only & $\mathbf{61.45 \pm 1.36}$ & $82.88 \pm 0.79$ & $\mathbf{85.45 \pm 0.19}$ & $\mathbf{61.16 \pm 1.79}$ & $\underline{83.38 \pm 0.79}$ & $\underline{86.86 \pm 0.61}$ \\
& Entropy only & $\underline{61.31 \pm 1.32}$ & $\mathbf{82.94 \pm 0.78}$ & $\underline{85.44 \pm 0.32}$ & $\underline{60.85 \pm 1.84}$ & $\mathbf{83.44 \pm 0.83}$ & $86.80 \pm 0.48$ \\
& Full BIDETA & $\mathbf{61.45 \pm 1.36}$ & $\mathbf{82.94 \pm 0.79}$ & $\mathbf{85.45 \pm 0.19}$ & $\mathbf{61.16 \pm 1.79}$ & $\underline{83.41 \pm 0.79}$ & $\mathbf{86.87 \pm 0.58}$ \\
\bottomrule
\end{tabular}
\end{adjustbox}
\end{table}

Table~\ref{tab:fullablationf1} reports Macro-F1 for 11 BIDETA variants on SITR. Spectral normalization reshapes the standardized feature geometry and consistently strengthens class-balanced discrimination, while the uncertainty gate regulates anchor and graph evidence to reduce unreliable propagation. Removing spectral normalization produces the largest decline among the full-model variants; separating the gate branches shows that disagreement matches the complete gate in several settings, whereas entropy contributes more strongly with Sparsh at the 5\% budget. Overall, the complete framework provides the strongest and most consistent performance pattern, indicating that memory construction, spectral geometry, and reliability-aware inference form a tightly coordinated adaptation process.

\section{Parameter sensitivity}
\label{app:sensitivity}
\subsection{Training-only inner validation on SITR}
Tables~\ref{tab:sens8}--\ref{tab:sens14} evaluate parameter sensitivity on Mini\_3 with a 5\% support budget. Selecting parameters on one training-only sensor separates configuration selection from the six-sensor evaluation and improves experimental fairness. The disagreement weight and iteration count remain comparatively stable around their selected values, whereas graph temperature and recurrence range are more sensitive because they directly control edge concentration and the strength of recurrent evidence propagation.
\begin{table}[!htbp]
\centering
\caption{\textbf{Neighborhood-size sensitivity on SITR.} Accuracy (\%) on the Mini\_3 inner validation split with 5\% support is reported as mean $\pm$ sample standard deviation over three random seeds. $^{*}$ indicates the selected value.}
\label{tab:sens8}
\small
\setlength{\tabcolsep}{3pt}
\begin{adjustbox}{max width=\linewidth}
\begin{tabular}{lcc}
\toprule
Value & TVL & Sparsh \\
\midrule
10 & $89.74 \pm 2.69$ & $87.25 \pm 3.67$ \\
20 & $91.17 \pm 2.11$ & $88.01 \pm 4.92$ \\
40 & $92.38 \pm 1.91$$^{*}$ & $88.39 \pm 4.82$$^{*}$ \\
80 & $92.06 \pm 2.28$ & $87.28 \pm 5.16$ \\
120 & $91.45 \pm 2.30$ & $85.49 \pm 4.26$ \\
160 & $90.57 \pm 2.37$ & $84.71 \pm 4.22$ \\
\bottomrule
\end{tabular}
\end{adjustbox}
\end{table}

\begin{table}[!htbp]
\centering
\caption{\textbf{Graph-temperature sensitivity on SITR.} Accuracy (\%) on the Mini\_3 inner validation split with 5\% support is reported as mean $\pm$ sample standard deviation over three random seeds. $^{*}$ indicates the selected value.}
\label{tab:sens9}
\small
\setlength{\tabcolsep}{3pt}
\begin{adjustbox}{max width=\linewidth}
\begin{tabular}{lcc}
\toprule
Value & TVL & Sparsh \\
\midrule
0.03 & $84.92 \pm 2.29$ & $86.03 \pm 4.05$ \\
0.05 & $86.52 \pm 2.72$ & $86.55 \pm 4.24$ \\
0.07 & $88.16 \pm 2.59$ & $86.68 \pm 4.88$ \\
0.1 & $90.00 \pm 2.48$ & $87.38 \pm 4.91$ \\
0.2 & $92.38 \pm 1.91$$^{*}$ & $88.39 \pm 4.82$$^{*}$ \\
\bottomrule
\end{tabular}
\end{adjustbox}
\end{table}

\begin{table}[!htbp]
\centering
\caption{\textbf{Spectral-exponent sensitivity on SITR.} Accuracy (\%) on the Mini\_3 inner validation split with 5\% support is reported as mean $\pm$ sample standard deviation over three random seeds. $^{*}$ indicates the selected value.}
\label{tab:sens10}
\small
\setlength{\tabcolsep}{3pt}
\begin{adjustbox}{max width=\linewidth}
\begin{tabular}{lcc}
\toprule
Value & TVL & Sparsh \\
\midrule
0.0 & $88.61 \pm 2.23$ & $83.66 \pm 5.34$ \\
0.05 & $89.40 \pm 2.07$ & $84.43 \pm 5.36$ \\
0.15 & $90.39 \pm 1.81$ & $85.59 \pm 5.41$ \\
0.25 & $91.25 \pm 1.46$ & $86.63 \pm 5.68$ \\
0.4 & $92.03 \pm 1.76$ & $87.77 \pm 5.21$ \\
0.6 & $92.38 \pm 1.91$$^{*}$ & $88.39 \pm 4.82$$^{*}$ \\
\bottomrule
\end{tabular}
\end{adjustbox}
\end{table}
 
\begin{table}[!htbp]
\centering
\caption{\textbf{Disagreement-weight sensitivity on SITR.} Accuracy (\%) on the Mini\_3 inner validation split with 5\% support is reported as mean $\pm$ sample standard deviation over three random seeds. $^{*}$ indicates the selected value.}
\label{tab:sens11}
\small
\setlength{\tabcolsep}{3pt}
\begin{adjustbox}{max width=\linewidth}
\begin{tabular}{lcc}
\toprule
Value & TVL & Sparsh \\
\midrule
0.0 & $92.38 \pm 1.94$ & $88.37 \pm 4.83$ \\
0.25 & $92.38 \pm 1.91$ & $88.36 \pm 4.84$ \\
0.5 & $92.38 \pm 1.91$$^{*}$ & $88.39 \pm 4.82$$^{*}$ \\
0.75 & $92.38 \pm 1.91$ & $88.35 \pm 4.86$ \\
1.0 & $92.37 \pm 1.89$ & $88.31 \pm 4.90$ \\
\bottomrule
\end{tabular}
\end{adjustbox}
\end{table}

\begin{table}[!htbp]
\centering
\caption{\textbf{Gate-exponent sensitivity on SITR.} Accuracy (\%) on the Mini\_3 inner validation split with 5\% support is reported as mean $\pm$ sample standard deviation over three random seeds. $^{*}$ indicates the selected value.}
\label{tab:sens12}
\small
\setlength{\tabcolsep}{3pt}
\begin{adjustbox}{max width=\linewidth}
\begin{tabular}{lcc}
\toprule
Value & TVL & Sparsh \\
\midrule
0.05 & $92.38 \pm 1.91$$^{*}$ & $88.39 \pm 4.82$$^{*}$ \\
0.15 & $92.06 \pm 2.05$ & $88.18 \pm 4.73$ \\
0.3 & $91.48 \pm 2.20$ & $87.97 \pm 4.64$ \\
0.5 & $91.09 \pm 2.37$ & $87.96 \pm 4.48$ \\
1.0 & $90.47 \pm 2.48$ & $87.51 \pm 4.25$ \\
\bottomrule
\end{tabular}
\end{adjustbox}
\end{table}

\begin{table}[!htbp]
\centering
\caption{\textbf{Recurrence-range sensitivity on SITR.} Accuracy (\%) on the Mini\_3 inner validation split with 5\% support is reported as mean $\pm$ sample standard deviation over three random seeds. $^{*}$ indicates the selected value.}
\label{tab:sens13}
\small
\setlength{\tabcolsep}{3pt}
\begin{adjustbox}{max width=\linewidth}
\begin{tabular}{lcc}
\toprule
Value & TVL & Sparsh \\
\midrule
0.20:0.40 & $85.17 \pm 2.07$ & $83.61 \pm 3.52$ \\
0.40:0.60 & $88.05 \pm 2.44$ & $85.51 \pm 3.88$ \\
0.50:0.70 & $89.51 \pm 2.51$ & $86.68 \pm 4.32$ \\
0.60:0.80 & $90.98 \pm 2.48$ & $87.81 \pm 4.70$ \\
0.70:0.90 & $92.38 \pm 1.91$$^{*}$ & $88.39 \pm 4.82$$^{*}$ \\
\bottomrule
\end{tabular}
\end{adjustbox}
\end{table}

\begin{table}[!htbp]
\centering
\caption{\textbf{Iteration-count sensitivity on SITR.} Accuracy (\%) on the Mini\_3 inner validation split with 5\% support is reported as mean $\pm$ sample standard deviation over three random seeds. $^{*}$ indicates the selected value.}
\label{tab:sens14}
\small
\setlength{\tabcolsep}{3pt}
\begin{adjustbox}{max width=\linewidth}
\begin{tabular}{lcc}
\toprule
Value & TVL & Sparsh \\
\midrule
1 & $88.24 \pm 2.88$ & $84.96 \pm 3.63$ \\
3 & $91.80 \pm 2.13$ & $87.67 \pm 4.74$ \\
5 & $92.38 \pm 1.91$ & $88.15 \pm 4.79$ \\
10 & $92.38 \pm 2.08$$^{*}$ & $88.39 \pm 4.82$$^{*}$ \\
20 & $92.21 \pm 2.33$ & $88.18 \pm 4.97$ \\
30 & $92.08 \pm 2.28$ & $87.57 \pm 4.02$ \\
50 & $91.91 \pm 2.20$ & $86.45 \pm 4.39$ \\
\bottomrule
\end{tabular}
\end{adjustbox}
\end{table}

\FloatBarrier

\subsection{TacQuad ranking sensitivity}
Tables~\ref{tab:sens15}--\ref{tab:sens19} report parameter sensitivity for identity ranking on TacQuad. At the default spectral exponent of 0.25, spectral normalization improves MRR over an exponent of zero across all six backbone--budget settings. Neighborhood size and disagreement weight remain comparatively stable, whereas the spectral exponent and recurrence length have larger effects because they reshape inter-identity neighborhoods and regulate how strongly ranking evidence is propagated; this influence is most visible under the lowest support budget.
\begin{table}[!htbp]
\centering
\caption{\textbf{Neighborhood-size sensitivity on TacQuad.} MRR (\%) for identity ranking is reported as mean $\pm$ sample standard deviation over three random seeds. $^{*}$ indicates the default value.}
\label{tab:sens15}
\small
\setlength{\tabcolsep}{3pt}
\begin{adjustbox}{max width=\linewidth}
\begin{tabular}{lcccccc}
\toprule
Value & TVL-10\% & TVL-20\% & TVL-30\% & Sparsh-10\% & Sparsh-20\% & Sparsh-30\% \\
\midrule
10 & $61.38 \pm 1.04$ & $77.14 \pm 0.30$ & $80.88 \pm 1.76$ & $69.61 \pm 0.60$ & $80.82 \pm 2.81$ & $86.04 \pm 1.45$ \\
20 & $61.23 \pm 0.91$ & $77.09 \pm 0.36$ & $81.10 \pm 1.84$ & $69.29 \pm 0.60$ & $80.86 \pm 2.71$ & $85.99 \pm 1.66$ \\
40 & $61.08 \pm 0.86$ & $76.97 \pm 0.19$ & $80.96 \pm 1.85$ & $69.26 \pm 0.66$ & $80.54 \pm 2.68$ & $85.96 \pm 1.71$ \\
80 $^*$ & $61.16 \pm 0.95$ & $76.98 \pm 0.20$ & $80.90 \pm 1.89$ & $69.28 \pm 0.72$ & $80.82 \pm 2.67$ & $85.86 \pm 1.80$ \\
120 & $61.17 \pm 0.95$ & $76.96 \pm 0.24$ & $80.85 \pm 1.93$ & $69.34 \pm 0.73$ & $80.83 \pm 2.63$ & $85.86 \pm 1.81$ \\
\bottomrule
\end{tabular}
\end{adjustbox}
\end{table}

\begin{table}[!htbp]
\centering
\caption{\textbf{Spectral-exponent sensitivity on TacQuad.} MRR (\%) for identity ranking is reported as mean $\pm$ sample standard deviation over three random seeds. $^{*}$ indicates the default value.}
\label{tab:sens16}
\small
\setlength{\tabcolsep}{3pt}
\begin{adjustbox}{max width=\linewidth}
\begin{tabular}{lcccccc}
\toprule
Value & TVL-10\% & TVL-20\% & TVL-30\% & Sparsh-10\% & Sparsh-20\% & Sparsh-30\% \\
\midrule
0.0 & $60.14 \pm 0.31$ & $75.63 \pm 0.74$ & $79.47 \pm 1.46$ & $68.22 \pm 0.53$ & $78.53 \pm 3.38$ & $84.00 \pm 2.17$ \\
0.1 & $60.64 \pm 0.50$ & $76.31 \pm 0.34$ & $80.22 \pm 1.67$ & $68.59 \pm 0.86$ & $79.44 \pm 3.20$ & $84.65 \pm 1.79$ \\
0.25 $^*$ & $61.16 \pm 0.95$ & $76.98 \pm 0.20$ & $80.90 \pm 1.89$ & $69.28 \pm 0.72$ & $80.82 \pm 2.67$ & $85.86 \pm 1.80$ \\
0.4 & $61.34 \pm 1.15$ & $77.26 \pm 0.10$ & $81.23 \pm 1.61$ & $69.47 \pm 0.66$ & $81.14 \pm 2.34$ & $86.22 \pm 1.49$ \\
0.6 & $61.61 \pm 1.10$ & $77.35 \pm 0.13$ & $81.52 \pm 1.72$ & $69.43 \pm 0.62$ & $81.21 \pm 2.31$ & $86.10 \pm 1.41$ \\
\bottomrule
\end{tabular}
\end{adjustbox}
\end{table}

\begin{table}[!htbp]
\centering
\caption{\textbf{Recurrence-range sensitivity on TacQuad.} MRR (\%) for identity ranking is reported as mean $\pm$ sample standard deviation over three random seeds. $^{*}$ indicates the default value.}
\label{tab:sens17}
\small
\setlength{\tabcolsep}{3pt}
\begin{adjustbox}{max width=\linewidth}
\begin{tabular}{lcccccc}
\toprule
Value & TVL-10\% & TVL-20\% & TVL-30\% & Sparsh-10\% & Sparsh-20\% & Sparsh-30\% \\
\midrule
0.20:0.40 & $60.88 \pm 0.41$ & $76.51 \pm 0.49$ & $80.55 \pm 1.09$ & $68.33 \pm 0.72$ & $80.07 \pm 2.70$ & $85.36 \pm 1.40$ \\
0.40:0.60 & $61.20 \pm 0.82$ & $76.83 \pm 0.35$ & $80.75 \pm 1.31$ & $68.86 \pm 0.92$ & $80.37 \pm 2.57$ & $85.57 \pm 1.72$ \\
0.60:0.80 $^*$ & $61.16 \pm 0.95$ & $76.98 \pm 0.20$ & $80.90 \pm 1.89$ & $69.28 \pm 0.72$ & $80.82 \pm 2.67$ & $85.86 \pm 1.80$ \\
0.70:0.90 & $61.06 \pm 1.01$ & $76.98 \pm 0.23$ & $80.62 \pm 1.93$ & $69.71 \pm 0.57$ & $80.95 \pm 2.81$ & $86.15 \pm 1.56$ \\
\bottomrule
\end{tabular}
\end{adjustbox}
\end{table}

\begin{table}[!htbp]
\centering
\caption{\textbf{Disagreement-weight sensitivity on TacQuad.} MRR (\%) for identity ranking is reported as mean $\pm$ sample standard deviation over three random seeds. $^{*}$ indicates the default value.}
\label{tab:sens18}
\small
\setlength{\tabcolsep}{3pt}
\begin{adjustbox}{max width=\linewidth}
\begin{tabular}{lcccccc}
\toprule
Value & TVL-10\% & TVL-20\% & TVL-30\% & Sparsh-10\% & Sparsh-20\% & Sparsh-30\% \\
\midrule
0.0 & $61.24 \pm 0.92$ & $76.96 \pm 0.34$ & $80.78 \pm 1.93$ & $69.16 \pm 0.75$ & $80.82 \pm 2.68$ & $85.85 \pm 1.77$ \\
0.25 & $61.21 \pm 0.93$ & $77.03 \pm 0.29$ & $80.85 \pm 1.91$ & $69.21 \pm 0.78$ & $80.84 \pm 2.71$ & $85.88 \pm 1.79$ \\
0.5 $^*$ & $61.16 \pm 0.95$ & $76.98 \pm 0.20$ & $80.90 \pm 1.89$ & $69.28 \pm 0.72$ & $80.82 \pm 2.67$ & $85.86 \pm 1.80$ \\
0.75 & $61.16 \pm 0.96$ & $76.97 \pm 0.17$ & $80.93 \pm 1.88$ & $69.30 \pm 0.70$ & $80.76 \pm 2.68$ & $85.91 \pm 1.75$ \\
1.0 & $61.10 \pm 0.92$ & $77.04 \pm 0.19$ & $80.94 \pm 1.88$ & $69.39 \pm 0.66$ & $80.77 \pm 2.69$ & $85.91 \pm 1.75$ \\
\bottomrule
\end{tabular}
\end{adjustbox}
\end{table}

\begin{table}[H]
\centering
\caption{\textbf{Iteration-count sensitivity on TacQuad.} MRR (\%) for identity ranking is reported as mean $\pm$ sample standard deviation over three random seeds. $^{*}$ indicates the default value.}
\label{tab:sens19}
\small
\setlength{\tabcolsep}{3pt}
\begin{adjustbox}{max width=\linewidth}
\begin{tabular}{lcccccc}
\toprule
Value & TVL-10\% & TVL-20\% & TVL-30\% & Sparsh-10\% & Sparsh-20\% & Sparsh-30\% \\
\midrule
1 & $63.35 \pm 1.77$ & $78.01 \pm 0.60$ & $81.93 \pm 2.32$ & $69.49 \pm 0.82$ & $81.20 \pm 2.54$ & $85.82 \pm 1.19$ \\
3 & $62.42 \pm 1.41$ & $77.87 \pm 0.84$ & $81.40 \pm 2.20$ & $69.41 \pm 1.07$ & $81.37 \pm 2.83$ & $86.02 \pm 1.28$ \\
5 & $62.00 \pm 1.09$ & $77.49 \pm 0.70$ & $80.97 \pm 1.86$ & $69.42 \pm 0.81$ & $81.22 \pm 2.62$ & $85.76 \pm 1.32$ \\
10 & $61.18 \pm 0.89$ & $77.10 \pm 0.27$ & $80.92 \pm 1.82$ & $69.29 \pm 0.78$ & $80.88 \pm 2.77$ & $85.95 \pm 1.76$ \\
20 $^*$ & $61.16 \pm 0.95$ & $76.98 \pm 0.20$ & $80.90 \pm 1.89$ & $69.28 \pm 0.72$ & $80.82 \pm 2.67$ & $85.86 \pm 1.80$ \\
30 & $61.14 \pm 0.90$ & $76.88 \pm 0.12$ & $80.79 \pm 1.94$ & $69.27 \pm 0.71$ & $80.77 \pm 2.66$ & $85.86 \pm 1.83$ \\
\bottomrule
\end{tabular}
\end{adjustbox}
\end{table}

\FloatBarrier

\section{Algorithm Cost Comparison}
\label{app:cost}
\begin{table}[!htbp]
\centering
\caption{\textbf{Target-sensor adaptation and inference costs on three tactile datasets.} Adaptation time (s) is reported as mean $\pm$ sample standard deviation over three random seeds. Query/s denotes the estimated end-to-end throughput including frozen-encoder inference. SITR and TacVerse use 10\% support, while TacQuad uses 30\% support. SITR-Calib is unavailable on TacVerse; its TacQuad counterpart is denoted SITR-Support. Best results are in bold and second-best results are underlined.}
\label{tab:costcomparison}
\scriptsize
\setlength{\tabcolsep}{2pt}
\begin{adjustbox}{max width=\linewidth}
\begin{tabular}{lllcccccc}
\toprule
& & & \multicolumn{2}{c}{SITR} & \multicolumn{2}{c}{TacVerse Shape} & \multicolumn{2}{c}{TacQuad} \\
\cmidrule(lr){4-5}\cmidrule(lr){6-7}\cmidrule(lr){8-9}
Backbone & Method & Mode & Adapt. (s) $\downarrow$ & Query/s $\uparrow$ & Adapt. (s) $\downarrow$ & Query/s $\uparrow$ & Adapt. (s) $\downarrow$ & Query/s $\uparrow$ \\
\midrule
TVL & Frozen backbone & I & $\mathbf{0.000 \pm 0.000}$ & $\mathbf{7{,}168.1}$ & $\mathbf{0.000 \pm 0.000}$ & $\mathbf{7{,}156.6}$ & $\mathbf{0.000 \pm 0.000}$ & $\mathbf{7{,}082.3}$ \\
TVL & Tip-Adapter & I & $0.033 \pm 0.009$ & $6{,}770.7$ & $0.007 \pm 0.003$ & $6{,}932.0$ & $0.007 \pm 0.001$ & $6{,}793.0$ \\
TVL & SimpleShot & I & $0.009 \pm 0.000$ & $7{,}026.5$ & $0.003 \pm 0.000$ & $7{,}016.2$ & $\underline{0.004 \pm 0.000}$ & $6{,}970.9$ \\
TVL & LaplacianShot & T & $\underline{0.008 \pm 0.000}$ & $5{,}256.2$ & $\underline{0.002 \pm 0.000}$ & $5{,}688.7$ & $\underline{0.004 \pm 0.000}$ & $4{,}697.8$ \\
TVL & SITR-Calib / SITR-Support & I & $12.268 \pm 0.807$ & $7{,}118.3$ & -- & -- & $25.269 \pm 2.325$ & $6{,}903.6$ \\
TVL & AnyTouch Match & I & $10.249 \pm 0.394$ & $\underline{7{,}130.8}$ & $8.183 \pm 1.489$ & $\underline{7{,}112.1}$ & $22.330 \pm 3.052$ & $\underline{6{,}989.5}$ \\
TVL & CTSRL CSM & I & $13.232 \pm 0.223$ & $7{,}115.8$ & $10.623 \pm 0.105$ & $7{,}092.2$ & $27.206 \pm 0.590$ & $6{,}880.3$ \\
TVL & BIDETA & T & $0.636 \pm 0.015$ & $6{,}056.0$ & $0.514 \pm 0.031$ & $5{,}835.1$ & $0.846 \pm 0.051$ & $4{,}581.3$ \\
\midrule
Sparsh & Frozen backbone & I & $\mathbf{0.000 \pm 0.000}$ & $\mathbf{4{,}965.2}$ & $\mathbf{0.000 \pm 0.000}$ & $\mathbf{4{,}958.5}$ & $\mathbf{0.000 \pm 0.000}$ & $\mathbf{4{,}930.4}$ \\
Sparsh & Tip-Adapter & I & $0.034 \pm 0.011$ & $4{,}845.3$ & $0.005 \pm 0.000$ & $4{,}884.9$ & $0.007 \pm 0.001$ & $4{,}830.3$ \\
Sparsh & SimpleShot & I & $\underline{0.009 \pm 0.001}$ & $4{,}891.2$ & $\underline{0.002 \pm 0.000}$ & $4{,}888.8$ & $\underline{0.004 \pm 0.000}$ & $4{,}866.6$ \\
Sparsh & LaplacianShot & T & $\underline{0.009 \pm 0.000}$ & $3{,}980.5$ & $\underline{0.002 \pm 0.000}$ & $4{,}167.6$ & $\underline{0.004 \pm 0.000}$ & $3{,}593.8$ \\
Sparsh & SITR-Calib / SITR-Support & I & $12.600 \pm 1.182$ & $4{,}943.1$ & -- & -- & $24.219 \pm 0.175$ & $4{,}840.1$ \\
Sparsh & AnyTouch Match & I & $10.316 \pm 0.110$ & $\underline{4{,}951.2}$ & $7.655 \pm 0.882$ & $\underline{4{,}942.2}$ & $23.770 \pm 0.570$ & $\underline{4{,}881.0}$ \\
Sparsh & CTSRL CSM & I & $13.688 \pm 0.155$ & $4{,}944.1$ & $10.136 \pm 0.659$ & $4{,}929.0$ & $24.917 \pm 1.456$ & $4{,}838.2$ \\
Sparsh & BIDETA & T & $0.283 \pm 0.034$ & $4{,}545.1$ & $0.160 \pm 0.001$ & $4{,}477.0$ & $0.293 \pm 0.015$ & $3{,}747.4$ \\
\bottomrule
\end{tabular}
\end{adjustbox}
\end{table}

\FloatBarrier

\end{document}